%% file: main.tex
\documentclass[10pt,twocolumn,letterpaper]{article}

\usepackage[pagenumbers]{cvpr} 
\usepackage[accsupp]{axessibility}
\definecolor{cvprblue}{rgb}{0.21,0.49,0.74}
\definecolor{Red}{RGB}{255, 1, 0}
\definecolor{OliveGreen}{rgb}{0.36, 0.71, 0.39}

\usepackage[pagebackref,breaklinks,colorlinks,allcolors=cvprblue]{hyperref}
\usepackage{amsmath}
\usepackage{amssymb}
\usepackage{booktabs}
\usepackage{tabularx}
\usepackage{multirow,multicol}
\usepackage{comment}
\usepackage[breakable]{tcolorbox}

\usepackage{microtype}
\def\vidhalluc{\textsc{VidHalluc}}

\def\paperID{17614} 
\def\confName{CVPR}
\def\confYear{2025}

\title{\vidhalluc: Evaluating Temporal Hallucinations\\
in Multimodal Large Language Models for Video Understanding}

\author{
Chaoyu Li
\qquad \qquad
Eun Woo Im
\qquad \qquad
Pooyan Fazli\\
Arizona State University\\
{\small\tt \{chaoyuli, eunwooim, pooyan\}@asu.edu}
}

\begin{document}
\maketitle
\input{sec/0_abstract}
\input{sec/1_intro}
\input{sec/2_related_works}
\input{sec/3_VidHalluc}
\input{sec/4_DINO-HEAL}
\input{sec/5_experiments}
\input{sec/6_conclusion}
\clearpage
{
    \section*{Acknowledgments}
    This research was supported by the National Eye Institute (NEI) of the National Institutes of Health (NIH) under award number R01EY034562.\ The content is solely the responsibility of the authors and does not necessarily represent the official views of the NIH. 
    \small  
    \bibliographystyle{ieeenat_fullname}
    \bibliography{main}
}
\input{sec/X_suppl}

\end{document}

%% file: sec/0_abstract.tex
\begin{abstract}
Multimodal large language models (MLLMs) have recently shown significant advancements in video understanding, excelling in content reasoning and instruction-following tasks. However, hallucination, where models generate inaccurate or misleading content, remains underexplored in the video domain.\ Building on the observation that MLLM visual encoders often fail to distinguish visually different yet semantically similar video pairs, we introduce \vidhalluc, the largest benchmark designed to examine hallucinations in MLLMs for video understanding.\ It consists of 5,002 videos, paired to highlight cases prone to hallucinations.\ \vidhalluc~assesses hallucinations across three critical dimensions: (1) action, (2) temporal sequence, and (3) scene transition. Comprehensive testing shows that most MLLMs are vulnerable to hallucinations across these dimensions. Furthermore, we propose DINO-HEAL, a training-free method that reduces hallucinations by incorporating spatial saliency from DINOv2 to reweight visual features during inference. Our results show that DINO-HEAL consistently improves performance on \vidhalluc, achieving an average improvement of 3.02\% in mitigating hallucinations across all tasks. Both the \vidhalluc~benchmark and DINO-HEAL code are available at \url{https://people-robots.github.io/vidhalluc}.

\end{abstract}
\vspace{-1em}

%% file: sec/1_intro.tex
\section{Introduction}
\label{sec:intro}

\begin{figure}
    \centering
    \includegraphics[width=0.48\textwidth]{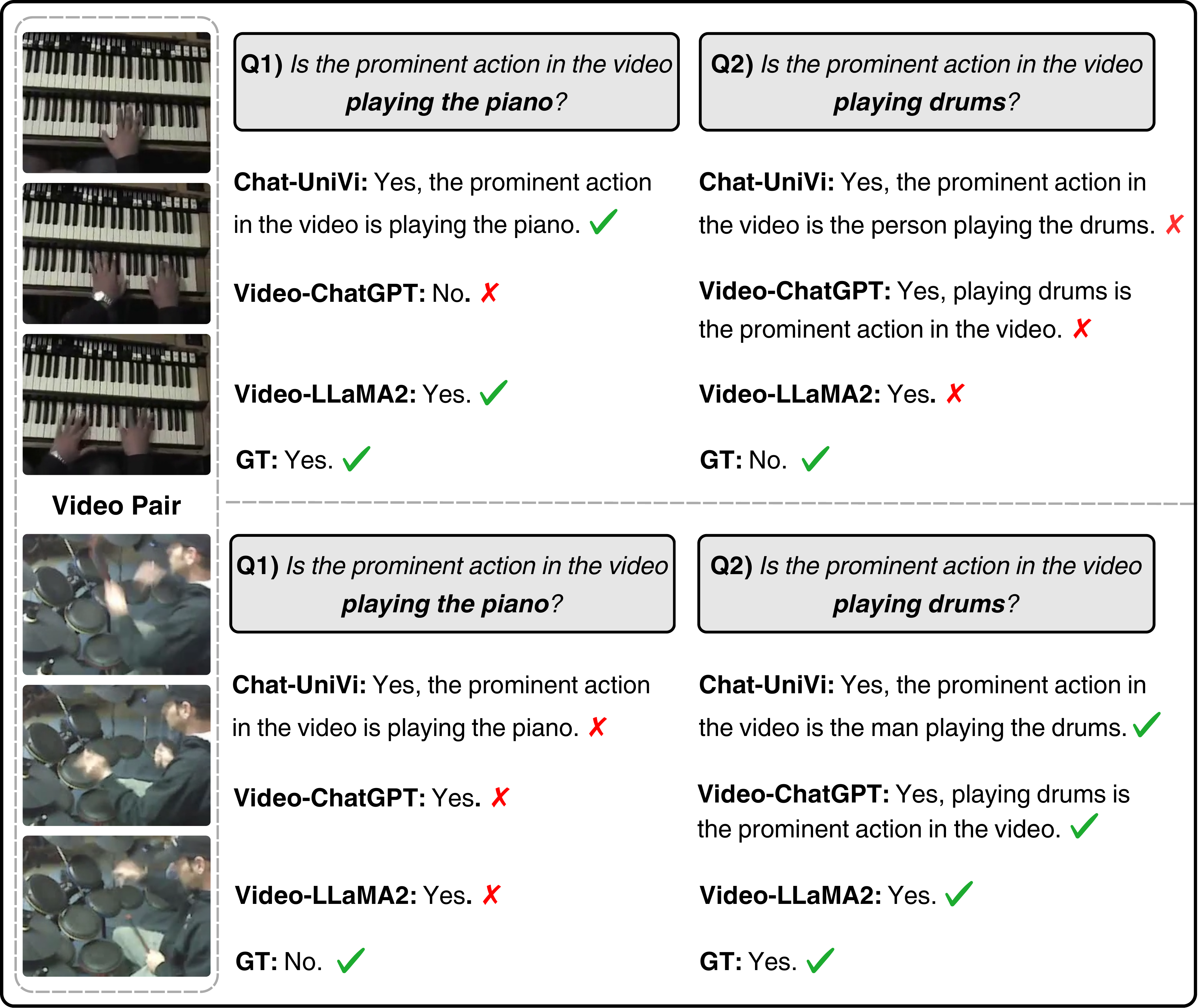}
    \caption{An example of a video pair showing action hallucination in \vidhalluc.\
   Adversarial questions, which refer to actions in the other video of the pair, show that MLLMs are prone to hallucinations when input videos have high semantic but low visual similarity. None of the models accurately identify the prominent actions in this pair.
    }
    \label{fig:samples}
    \vspace{-1.5em}
\end{figure}



The rapid advancement of multimodal large language models (MLLMs)~\citep{cambrian,llava, videollava} has led to significant improvements in video understanding, particularly in semantic reasoning and instruction-following~\cite{mllm_survey}. Despite these improvements, MLLMs tend to generate hallucinations, producing plausible but factually incorrect information~\cite{vcd}, raising concerns about their reliability in practical applications.\ To better understand and quantify these limitations, several benchmarks~\cite{videohallucer, pope, hallusionbench} and mitigation strategies~\cite{ha_dpo, hallucidoctor} have been developed to systematically evaluate hallucinations in MLLMs.\
These benchmarks curate fine-grained question-answer (QA) pairs designed to trigger hallucinations, often created using either GPT-4V~\cite{sharegpt4video, vript} or human annotators~\cite{mangalam2023egoschema}.
However, these benchmarks come with several limitations.
First, the datasets are typically small in scale, containing fewer than 1,000 videos and 2,000 QA pairs.
This limited scale is due to the time and cost involved in generating high-quality QA pairs, making the process inefficient.
Second, the primary focus of the existing benchmarks is on hallucinations in static elements of videos~\citep{mhalubench, phd}, such as objects~\cite{pope, rohrbach2018object}, their attributes~\cite{mmhal_bench, amber}, and spatial relationships~\cite{han2024instinctive}.
This limits the benchmark's efficacy in evaluating dynamic and temporal content in videos.
Third, the benchmarks are constructed with limited question types (e.g., binary QA only), constraining the assessment of the model's understanding of dynamic aspects, such as temporal scene transitions, inherent to video content.

To address these gaps, we introduce \vidhalluc, the largest benchmark for evaluating hallucinations in video understanding, comprising 5,002 videos and 9,295 QA pairs. To curate the benchmark, we build on prior work \cite{mmvp} to develop an automated data collection pipeline that generates potential hallucination video pairs based on semantic similarity and visual differences. Moreover, while previous benchmarks often focus on static tasks, \vidhalluc~specifically targets dynamic elements of videos that cause hallucinations, namely, action, temporal sequence, and scene transition hallucinations.

On the other hand, efforts have been made to mitigate MLLM hallucinations in video understanding. These methods include constructing high-quality instruction-tuning datasets~\cite{lrv-instruction}, enhancing input resolution~\cite{liu2024improved}, and leveraging reinforcement learning from human feedback (RLHF)~\cite{ha_dpo, zhou2024aligning, mmhal_bench, yu2024rlhf}.
However, these data-centric optimization approaches still require fine-tuning, which incurs significant computational costs and necessitates creating additional datasets.
As shown in Figure~\ref{fig:samples}, we observe that current MLLMs are vulnerable to hallucinations when they encounter semantically similar videos.
We attribute these hallucinations to the inductive bias inherent in the visual encoder (e.g., the CLIP series), which emphasizes the discrimination of contextual scenes due to its image-text contrastive learning. This creates a discrepancy between the encoder's provision of semantic information about the video and the language model’s need for both semantic and vision-only unimodal representations~\cite{mmvp}.

To address this issue, we propose DINO-HEAL, a novel, training-free algorithm designed to mitigate MLLM hallucinations in video understanding by enhancing the visual encoder's focus on salient spatial regions during inference.
DINO-HEAL applies to any CLIP-series vision encoder.\
To reduce reliance on the CLIP series and better encode spatially important features, DINO-HEAL uses saliency maps from DINOv2~\cite{dinov2} to reweight frame features, highlighting key regions.\
This approach enables models to achieve substantial accuracy gains, such as +5.01\% for Video-LLaVA~\cite{videollava} and +4.77\% for Video-ChatGPT~\cite{videochatgpt} in action hallucination, as well as notable improvements in temporal sequence hallucination, with Video-ChatGPT and VideoLLaMA2~\cite{videollama2} achieving gains of +11.67\% and +18.83\%, respectively.

In summary, our main contributions are as follows:

\begin{itemize}
    \item We introduce \vidhalluc, the largest benchmark for assessing hallucinations in MLLMs for video understanding, designed to evaluate action, temporal sequence, and scene transition hallucinations.
    \item We develop a training-free algorithm, DINO-HEAL, to enhance the visual encoder's focus on critical regions and improve the model's robustness against hallucinations.
    \item We conduct extensive experiments on \vidhalluc~with ten state-of-the-art models and provide a comprehensive analysis of the benchmark results.
    Additionally, we demonstrate the effectiveness of DINO-HEAL, achieving an average gain of 3.02\% across all hallucination categories and five models.
\end{itemize}

\begin{table*}[htbp]
    \centering
    \resizebox{\textwidth}{!}{
    \begin{tabular}{l c c c c c c c}
        \toprule
        Benchmark & \# of Ques. / \# of Videos & Binary QA & MCQ & Open-Ended Ques. & Sorting Ques. & Control Pairs & Adversarial\\
        \midrule
        HallusionBench~\cite{hallusionbench} & 1129 / 346 & \checkmark & \checkmark & \texttimes & \texttimes & \checkmark & \checkmark\\
        VideoHallucer~\cite{videohallucer} & 1,800 / 948 & \checkmark & \texttimes & \texttimes & \texttimes& \texttimes & \checkmark\\
        Vript-HAL~\cite{vript} & 122 / 122 & \texttimes & \texttimes & \checkmark & \texttimes & \texttimes & \texttimes \\
        EventHallusion~\cite{eventhallusion} & - / 400 & \checkmark & \texttimes & \checkmark & \texttimes & \texttimes & \texttimes \\
        \midrule
        \textbf{\vidhalluc~(Ours)} & 9,295 / 5,002 & \checkmark & \checkmark & \checkmark & \checkmark & \checkmark & \checkmark \\
        \bottomrule
    \end{tabular}
    }
    \caption{Comparison of \vidhalluc~with recent hallucination benchmarks in video understanding. \vidhalluc~is the largest benchmark for evaluating hallucinations, supporting diverse question formats, including binary QA, MCQ, open-ended, and sorting questions, with control pairs and adversarial evaluation to enhance robustness.}
    \label{tab:benchmark_comparison}
    \vspace{-1em}
\end{table*}

%% file: sec/2_related_works.tex
\section{Related Works}
\label{sec:related_works}

\subsection{Multimodal Large Language Models}

Building on the success of large language models (LLMs)~\cite{devlin2018bert, gpt3, wei2021finetuned, llama, palm-e, t5, vicuna, chung2024scaling}, researchers have enhanced these models with vision capabilities, leading to the development of MLLMs~\cite{blip2, instructblip, minigpt}. This is achieved by processing images as tokens concatenated with text tokens through cross-attention~\cite{flamingo} or by directly aligning visual features with nonlinear projections~\cite{llava, fromage}. Recent advancements have led to numerous MLLMs designed specifically for video understanding~\cite{chatunivi, vila}, which build upon image approaches to address the sequential and dynamic nature of video content. For example, Video-ChatGPT~\cite{videochatgpt} employs temporal aggregation techniques to integrate information across frames, Video-LLaVA~\cite{videollava} uses time-sensitive attention mechanisms to manage frame dependencies, and LLaVA-NeXT-Video~\cite{llavanextvideo} introduces a linear scaling approach to process longer video sequences, effectively capturing extended temporal dependencies.


Most MLLMs use CLIP-based visual encoders~\cite{clip, siglip, evaclip, cherti2023reproducible, naeem2025silc, liu2024improved} for robust visual-text alignment. However, these encoders, shaped by image-text contrastive learning, struggle with dynamic scenes and temporal relationships in videos~\cite{videoclip}. To assess and highlight these biases, \vidhalluc~ leverages CLIP/SigLIP and DINOv2~\cite{dinov2}, trained solely on image data to provide visual features, to curate video pairs that may cause confusion and hallucinations, offering a rigorous evaluation framework for MLLMs.

\subsection{Hallucination in MLLMs}

Hallucination in MLLMs occurs when the model generates inaccurate or entirely fabricated responses, failing to accurately reflect the input data.\ To mitigate this issue in image-based tasks, Woodpecker~\cite{woodpecker} uses post-hoc corrections through additional visual cues, while LRV-Instruction~\cite{lrv-instruction} and HalluciDoctor~\cite{hallucidoctor} offer balanced instruction-tuning datasets to reduce hallucinations. As MLLMs extend into video understanding tasks, new challenges arise in handling temporal inconsistencies.\ To tackle this, Volcano\cite{volcano} uses a critique-revise-decide framework for self-correction, and Vript~\cite{vript} enhances alignment by incorporating transcribed voice-overs, enriching multimodal representations. These methods showcase tailored strategies for minimizing hallucinations across different modalities in MLLMs.

While most methods require retraining or fine-tuning the MLLM to reduce hallucinations, our DINO-HEAL approach can be applied directly during inference without additional training, offering an efficient solution for mitigating hallucinations in resource-constrained scenarios.

\subsection{MLLM Benchmarks for Video Understanding}

As MLLMs continue to advance, conducting quantitative evaluations remains essential.\ Numerous efforts~\citep{oscar, mmbench, value, verified, vript, videovista} have been undertaken to assess various aspects of MLLMs in video understanding, encompassing tasks such as video reasoning~\cite{videovista}, text-to-video retrieval~\cite{value}, video captioning~\cite{verified}, and hallucination~\cite{vript}. These benchmarks assess models' capabilities in understanding and generating language from video content.\ For example, MVBench~\cite{mvbench} provides an extensive evaluation of temporal understanding in video-language models, encompassing 20 challenging video tasks that require dynamic analysis beyond individual frames.\ Video-MME~\cite{videomme} offers a comprehensive assessment of MLLMs' video analysis performance, covering diverse types of videos, varying durations, and multiple data modalities. In long-form video understanding, benchmarks such as HourVideo~\cite{hourvideo} and LongVideoBench~\cite{longvideobench} focus on event ordering.\
In the context of hallucination, HallusionBench~\cite{hallusionbench} reverses the frames in four-frame video clips and prompts the model to determine whether the events in the original captions align with those in the reversed video. 
Similarly, VideoHallucer~\cite{videohallucer} uses extended videos with multiple events to query the model on the correct sequence of these events. However, these hallucination benchmarks have limitations in scale, providing no more than 1,000 videos, which restricts their ability to comprehensively evaluate hallucinations.

%% file: sec/3_VidHalluc.tex
\begin{figure*}[h]
    \centering
    \includegraphics[width=1\textwidth]{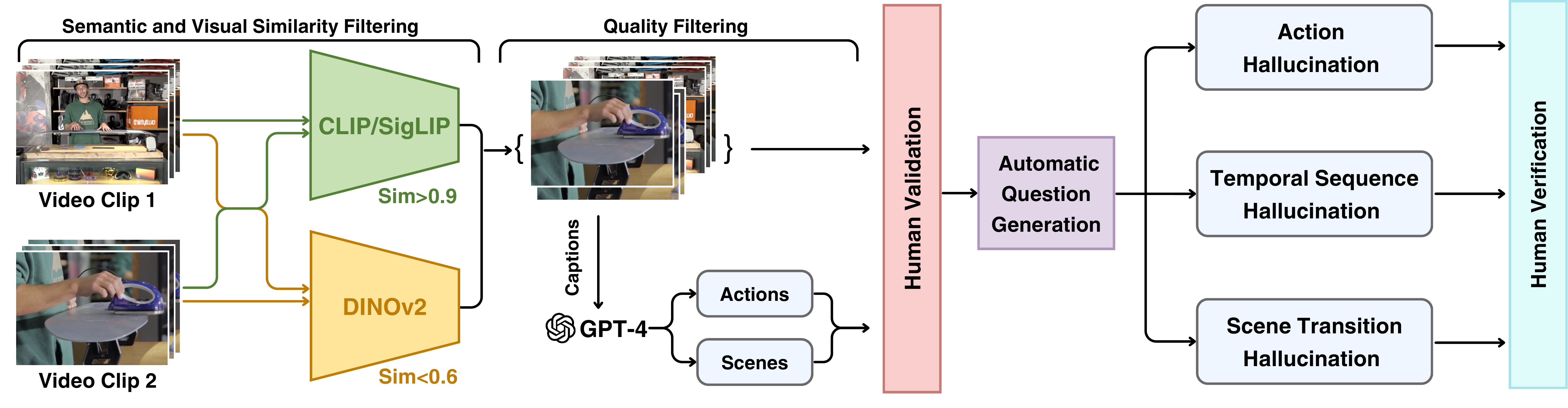}
    \caption{Overview of the \vidhalluc~benchmark construction process. Candidate video pairs are selected based on high semantic similarity and low visual similarity. GPT-4 is then used to generate action and scene annotations from the captions of each video clip. Human reviewers manually filter out pairs where GPT-4 annotations were incorrect or where actions/scenes did not align between clips. Finally, video pairs that pass this filtering process are used to automatically generate three types of hallucination questions: action hallucination, time-sequence hallucination, and scene-transition hallucination.}
    \label{fig:pipeline_benchmark2}
    \vspace{-1em}
\end{figure*}

\section{\vidhalluc~Benchmark}

\subsection{Data Collection}

Unlike other benchmarks for evaluating hallucinations in video understanding that rely on manually collected videos, our data collection pipeline is semi-automated (Figure~\ref{fig:pipeline_benchmark2}). To create the \vidhalluc~benchmark, we apply this pipeline to existing video description datasets, including ActivityNet \cite{activitynet}, YouCook2~\cite{youcook2}, and VALOR32K~\cite{valor}. In contrast to existing approaches, which typically use individual videos, our method treats video pairs as the primary data units. Below, we outline the key stages of our data collection process, including filtering for semantic and visual similarity, performing quality checks, conducting human validation, and generating hallucination-specific questions. For implementation details, please refer to the Supplementary Material.

\noindent\textbf{Semantic and Visual Similarity Filtering.} 
To identify potential hallucination pairs, we measure semantic similarity using CLIP/SigLIP and visual similarity with DINOv2. For semantic similarity, we apply average pooling to the features of CLIP/SigLIP, while DINOv2 uses a similar pooling method to calculate visual similarity. Video pairs scoring above the semantic threshold $\lambda_\text{sem}$ with CLIP/SigLIP but below the visual threshold $\lambda_\text{vis}$ with DINOv2 are flagged as hallucination candidates. These pairs highlight cases where models may over-rely on semantic cues, leading to potential misinterpretations of visually unique content. We use CLIP/SigLIP and DINOv2 for their complementary strengths: 
CLIP/SigLIP, trained on the cosine similarity between visual and textual projections, captures deep contextual relationships and thematic elements, while DINOv2, trained with a self-supervised objective, focuses on visual resemblance. This method also tests the model’s sensitivity to subtle visual differences that might be overlooked in single-video evaluations~\cite{mmvp}. $\lambda_{\text{sem}}$ and $\lambda_{\text{vis}}$ are set to 0.9 and 0.6, respectively.




\noindent\textbf{Quality Filtering.} After collecting the video pairs, we perform additional filtering to ensure dataset quality.\ First, we exclude pairs where either video is shorter than one second. Next, we leverage GPT-4~\cite{achiam2023gpt} to align each video with the actions and scenes described in the original dataset’s annotated captions. Finally, we filter out video pairs that contain identical extracted actions.

\noindent\textbf{Human Validation of Video Pairs.} We recruit four participants to conduct a manual review to eliminate pairs with the following issues: (1) the lack of a clear action in either video, (2) the presence of multiple actions in either video, or (3) identical actions in both videos. These rigorous steps result in a high-quality dataset, which serves as a reliable benchmark.

\begin{figure}
    \centering
    \includegraphics[width=0.5\textwidth]{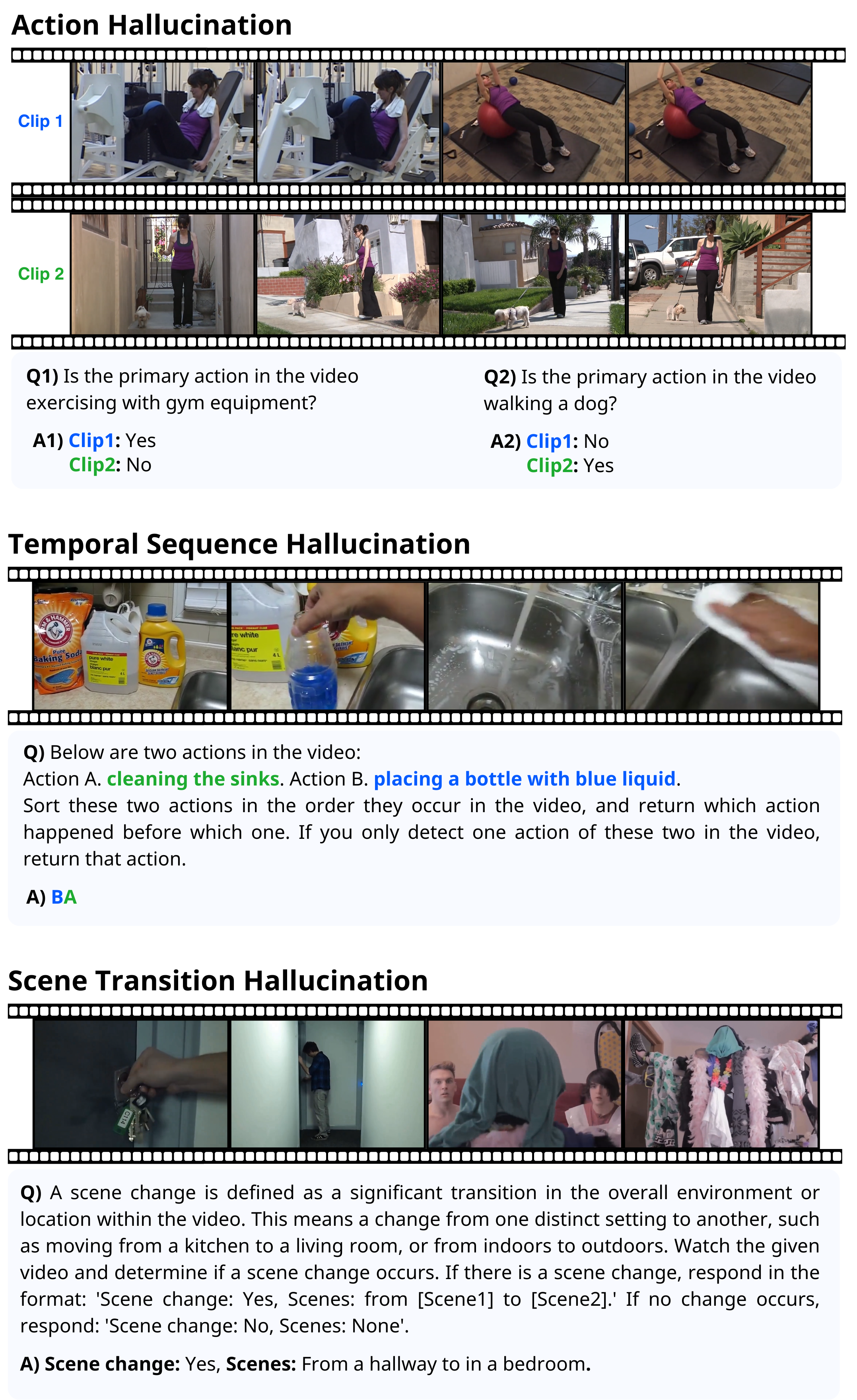}
    \caption{Examples of three hallucination types in the \vidhalluc~benchmark: (1) action hallucination, where the model detects actions in a video that significantly differ from the actual actions; (2) temporal sequence hallucination, where the model fails to represent the correct temporal order of events in a video; and (3) scene transition hallucination, where the model inaccurately describes transitions between distinct scenes within a video.}
    \label{fig:evaluation_categories}
    \vspace{-1.3em}
\end{figure}

\noindent\textbf{Automatic Question Generation.}\
We categorize the benchmark into three distinct hallucination types: (1) action hallucination, 
(2) temporal sequence hallucination, 
and (3) scene transition hallucination. 
For each type, we design specific question formats to evaluate model performance: binary QA and multiple-choice questions (MCQs) for action hallucination, sorting questions for temporal sequence hallucination, and open-ended questions for scene transition hallucination. After the filtering steps, the high-quality video pairs, along with their corresponding annotations, are further processed to automatically generate questions tailored to each hallucination type and question format. An example question for each hallucination type is shown in Figure~\ref{fig:evaluation_categories}.

\noindent\textbf{Human Verification of QAs.} We recruit nine participants to verify and refine the entire dataset. For ACH, they check whether the action described in the question matches the video and correct any inaccuracies. For TSH, they ensure that the action order in the answer aligns with the video, making revisions where necessary. For STH, they verify that the scene described in the answer corresponds to the video and adjust any mismatches. Following verification, 939 ACH, 68 TSH, and 38 STH questions are modified, resulting in an accuracy of 88.76\% for the automatically generated QAs.


\subsection{Action Hallucination}

Action hallucination (ACH) occurs when a model identifies actions in a video that are either non-existent or significantly different from the actual actions. This typically arises when the model misinterprets visual cues or overgeneralizes, leading to inaccurate conclusions about the video's content.\ In \vidhalluc, we evaluate the performance of MLLMs in detecting action hallucinations using two methods: binary QA and MCQs. The MCQs test the model's basic understanding of actions, assessing its ability to identify them without extra context. In contrast, binary QA evaluates the model’s susceptibility to hallucinations, especially when confronted with leading or suggestive questions.


For the binary QA, we use a question format such as ``\textit{Is the prominent action in the video \{action\}?''}. Each video pair involves two distinct actions. For instance, we first ask the question with Video 1, then with Video 2. The model should answer ``Yes'' for the video containing the action and ``No'' for the video without it. A correct response requires the model to answer both questions correctly. We evaluate the model's performance using accuracy, as defined in formula~\ref{eq:accuracy}, where $N_{\text{correct}}$ is the number of correctly answered questions and $N_{\text{total}}$ is the total number of questions.

\begin{equation}
\text{Accuracy} = N_{\text{correct}} / N_{\text{total}}.
\label{eq:accuracy} 
\end{equation}

\noindent For MCQs, we use a format such as ``\textit{What is the prominent action in the video?}''.\ Each question has four answer choices, with only one correct answer representing the action occurring in the video.\ One incorrect option corresponds to the action in the other video in the pair, while the remaining two options are generated using GPT-4o~\cite{openai2024gpt4ocard}.\ By providing a frame from the video, we ask GPT-4o to generate plausible actions that could occur in the same scene, but distinct from the correct action. The model must select the correct action from the four options. Accuracy remains the evaluation metric, calculated in the same manner as in Equation~\ref{eq:accuracy}.

\subsection{Temporal Sequence Hallucination}


Temporal sequence hallucination (TSH) occurs when a model misinterprets the order or timing of events in a video, confusing the sequence in which actions or scenes unfold. In \vidhalluc, model performance is evaluated on TSH by testing its ability to identify the correct sequence of events. To do this, we first select two video clips from a pair that appear consecutively in the original video. These clips are then concatenated in their original chronological order to form a longer video segment. The model’s task is to determine the correct sequence of actions within this concatenated segment. We use accuracy, calculated as in Equation~\ref{eq:accuracy}, as the evaluation metric to assess how well the model interprets the temporal relationships between the consecutive events.

\subsection{Scene Transition Hallucination}

Scene transition hallucination (STH) occurs when a model fails to detect or describe transitions between different scenes in a video. This can result in blending elements from different scenes or failing to recognize the start of a new scene, leading to inaccurate descriptions. We design evaluation tasks to assess MLLMs' ability to detect scene transitions accurately. To create these tasks, we first filter video pairs to identify those with distinct scenes, using captions from the original datasets to detect scene differences.
After filtering, each video pair is concatenated in both possible orders to produce two unique long video clips.
Additionally, an equal number of video pairs with no scene changes are collected to ensure dataset balance.
For video clips containing a scene transition, the ground truth labels indicate ``Yes'' for scene change and describe the transition as ``\textit{from [Scene1] to [Scene2]}''.
For clips without a scene change, the ground truth labels indicate ``No''.


The model is tasked with determining if a scene change occurs in each concatenated video segment and, if so, identifying the specific transition. We evaluate the model’s performance based on two criteria: (1) scene change classification, which assesses whether the model correctly identifies a scene change, and (2) scene transition description, which measures the model's ability to describe the transition from one scene to another accurately. For the scene change classification task, we calculate the Matthews correlation coefficient (MCC) between the model’s predictions and the ground truth labels, as defined in Equation~\ref{eq:mcc}:

\begin{equation}
   \frac{n_{11} \times n_{10} - n_{01} \times n_{00}}{\sqrt{(n_{11} + n_{01}) (n_{11} + n_{00})(n_{10} + n_{01})(n_{10} + n_{00})}},
    \label{eq:mcc} 
\end{equation}
\noindent where $A \in$ \{0 (False), 1 (True)\}, $P \in$ \{0 (Negative), 1 (Positive)\}, and $n_{AP}$ denote actual condition, predicted condition, and non-negative counts, respectively.
To adjust MCC to range between 0 and 1 and to further penalize models that consistently answer only ``Yes'' or only ``No'', we apply the transformation in Equation~\ref{eq:cls_score} to obtain the classification score $\text{Score}_{\text{cls}}$:
\begin{equation}
    \text{Score}_\text{cls} = \left( \frac{\text{MCC} + 1}{2} \right)^2.
    \label{eq:cls_score}
\end{equation}

\noindent The description task evaluates the model's ability to accurately recognize and describe scene information. To assess this, we extract scene descriptions from both the model’s output and the ground truth, formatting them for direct comparison. We then calculate the cosine similarity $S$ between the SimCSE~\cite{simcse} embeddings of the corresponding scenes. Based on this similarity measure, each scene description is assigned a score using Equation~\ref{eq:desc_score}:

\begin{equation}
    \text{Score}_{\text{desc}} = 
    \begin{cases} 
        \hfill 0, & \text{if } S \leq \text{THR}_{\text{low}} \\
        \frac{\sigma(S) - \sigma(\text{THR}_{\text{low}})}{\sigma(1) - \sigma(\text{THR}_{\text{low}})}, & \text{if } S > \text{THR}_{\text{low}}
    \end{cases},
    \label{eq:desc_score}
\end{equation}
where $S$ is the cosine similarity between the SimCSE~\cite{simcse} embeddings of the corresponding scenes, \( \text{THR}_{\text{low}} \) is the lowest threshold we set to start giving a score, $\sigma$ represents the Sigmoid function. The total description score is the average of the scores for the ``from'' and ``to'' scenes.

Finally, we calculate the overall evaluation score as a weighted sum of the classification score and the normalized description score as below:
\begin{equation}
    \text{Score}_{\text{overall}} = \alpha \times \text{Score}_{\text{cls}} + (1 - \alpha) \times \text{Score}_{\text{desc}}.
    \label{eq:overall_score}
\end{equation}

\begin{figure}
    \centering
    \includegraphics[width=\linewidth]{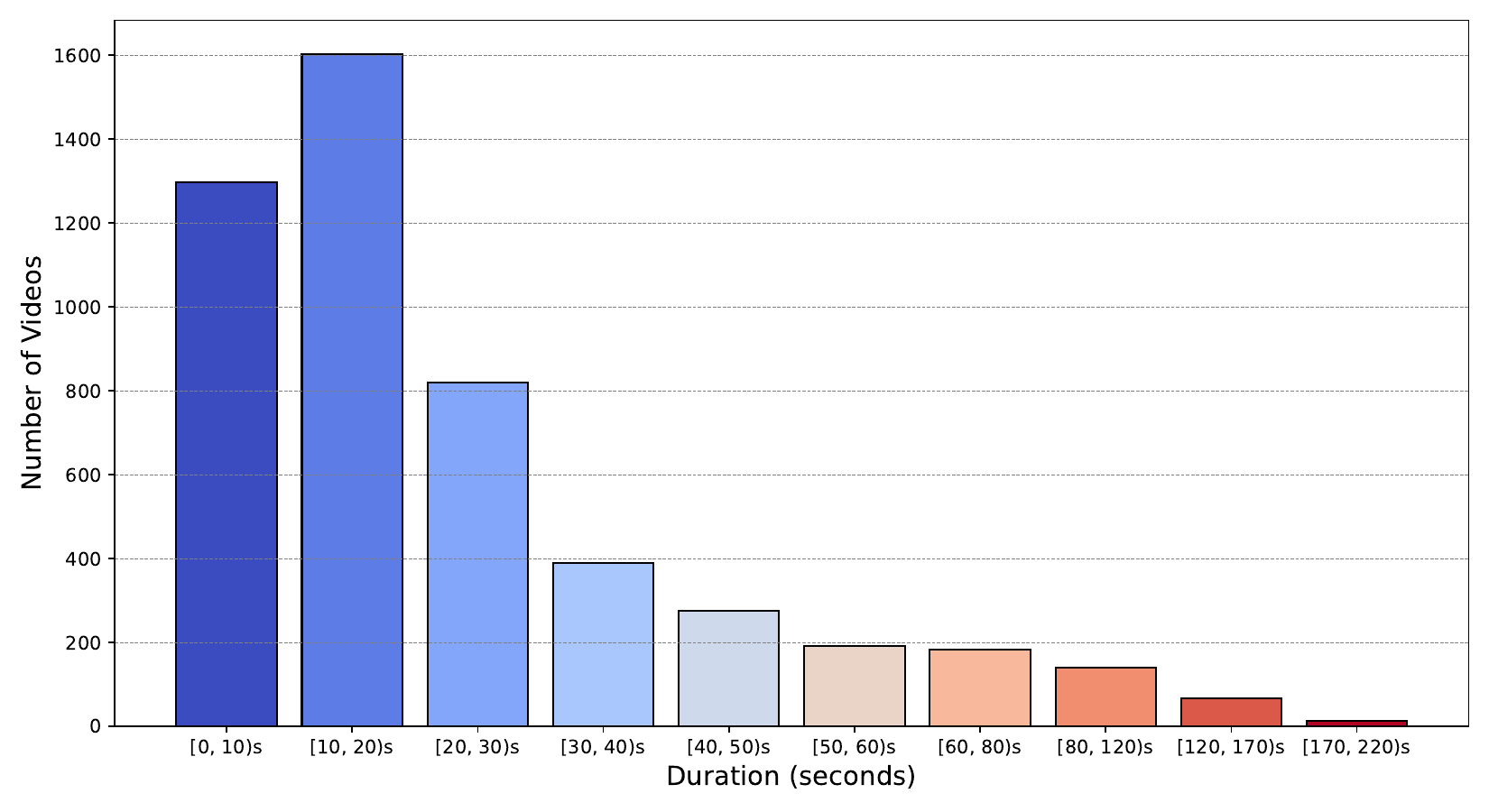}
    \vspace{-2em}
    \caption{The distribution of video durations in the \vidhalluc~benchmark.}
    \label{fig:video_duration}
    \vspace{-1.2em}
\end{figure}

\subsection{Dataset Statistics}


\noindent Table~\ref{tab:dataset_statistic} summarizes the statistics for \vidhalluc.\ The dataset consists of 5,002 unique videos in total, with 3,957 videos for ACH, 600 for TSH, and 445 for STH. \vidhalluc~also includes 9,295 QA pairs, with 8,250 questions for ACH, 600 for TSH, and 445 for STH. The average video length varies by hallucination type:
ACH videos have an average length of approximately 21.79 seconds, TSH videos average around 41.19 seconds, and STH videos average 28.72 seconds. The overall average length across the benchmark is 24.70 seconds. Figure~\ref{fig:video_duration} illustrates the distribution of video durations, showing a peak in the 10--20 second range. These statistics highlight the diversity and scale of \vidhalluc, establishing it as a valuable resource for evaluating video-based hallucinations.

\begin{table}[h]
    \centering
    \resizebox{0.45\textwidth}{!}{ 
    \begin{tabular}{l c c c c c c}
        \toprule
        Statistics & ACH & TSH & STH & Total \\
        \midrule
        \# of Videos & 3,957 & 600 & 445 & 5,002 \\
        \# of Questions & 8,250 & 600 & 445 & 9,295 \\
        Average Duration (s) & 21.79 & 41.19 & 28.72 & 24.70 \\
        \bottomrule
    \end{tabular}
    }
    \caption{\vidhalluc~statistics.}
    \label{tab:dataset_statistic}
    \vspace{-1.8em}
\end{table}



%% file: sec/4_DINO-HEAL.tex
\section{DINO-HEAL}


\begin{figure}
    \centering
    \includegraphics[width=0.5\textwidth]{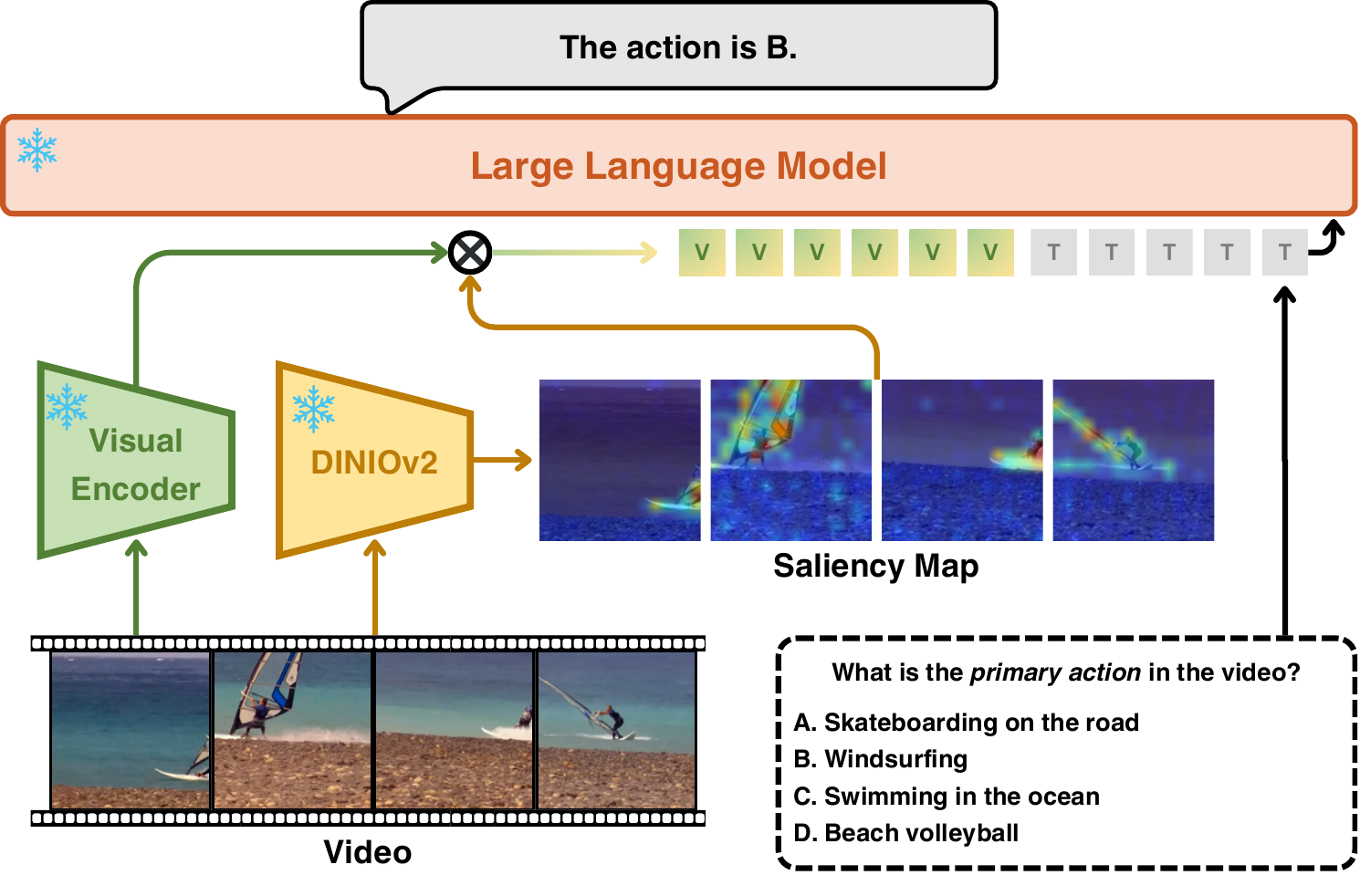}
    \caption{DINO-HEAL pipeline. Since DINOv2 effectively captures salient regions in the input video, we leverage it to guide the reweighting of the attention given to different spatial regions within the feature from the visual encoder.}
    \label{fig:pipeline_method}
    \vspace{-1em}
\end{figure}

We introduce DINO-HEAL, a training-free method that mitigates hallucinations by using saliency maps from DINOv2 to reweight features from the frozen visual encoder, emphasizing key spatial regions. DINO-HEAL requires no architectural modifications or additional training.

\subsection{DINOv2 Saliency Map Extraction}

For each input frame, the method processes it through the DINOv2 model to extract attention weights from the last layer and the $h$-th head, denoted as \( A_{\text{DINOv2}}^{(h)} \in \mathrm{R}^{L \times L} \), where $L$ is the sequence length (number of tokens), including the [CLS] token. We then compute the average attention across all heads:
\begin{equation}
A_{\text{DINOv2}} = \frac{1}{H} \sum_{h=1}^{H} A_{\text{DINOv2}}^{(h)},
\label{eq:avg_attention_weights}
\end{equation} 

\noindent where $H$ is the total number of attention heads. 
Then, we extract the attention values of the [CLS] token over all spatial tokens, excluding itself, as:
\begin{equation}
S_{\text{DINOv2}} = A_{\text{DINOv2}}[\text{CLS}, 1:L-1],
\label{eq:saliency_map_extraction}
\end{equation} 
where $S_{\text{DINOv2}}$ represents the saliency scores derived from DINOv2 for spatial tokens. We compute the saliency score for each spatial token by summing the attention weights across the query dimension:
\begin{equation} 
S_{\text{DINOv2}} = A_{\text{DINOv2}}^{\text{spatial}} \in \mathbb{R}^{L-1} .
\end{equation}
This results in a saliency map $S_{\text{DINOv2}}$ that represents the importance of each spatial token in the frame.

\begin{table*}[ht]
    \centering
    \resizebox{\textwidth}{!}{
    \begin{tabular}{lcccccccc}
        \toprule
        \multirow{2.5}{*}{\makebox[0.15\linewidth][c]{\textbf{Models}}} & \multirow{2.5}{*}{\centering \textbf{LLM Params}} & \multirow{2.5}{*}{\centering \textbf{Encoder}} & \multirow{2.5}{*}{\centering \textbf{Frame}} & \multicolumn{2}{c}{\textbf{Accuracy on ACH}} & \multirow{2.5}{*}{\textbf{Accuracy on TSH}$^\uparrow$} & \multirow{2.5}{*}{\textbf{Score on STH}$^\uparrow$} \\
        \cmidrule(r){5-6}
         & & & & \textbf{Binary QA}$^\uparrow$ & \textbf{MCQ}$^\uparrow$ & & \\
        \midrule
        Video-ChatGPT~\cite{videochatgpt} & 7B & CLIP ViT-L/14-224 & 100 & 9.50 & 24.58 & 30.17 & 7.70 \\
        Video-LLaVA~\cite{videollava} & 7B & LanguageBind-Video & 8 & 26.84 & 64.45 & 27.17 & 29.60 \\
        ShareGPT4Video~\cite{sharegpt4video} & 8B & CLIP ViT-L/14-336 & 16 & 29.96 & 44.78 & \underline{49.50} & 17.83\\
        Chat-UniVi~\cite{chatunivi} & 13B & CLIP ViT-L/14-224 & fps=1 & 23.77 & 54.79 & 35.50 & 29.87\\
        LLaVA-NeXT-Video~\cite{llavanextvideo} & 34B & CLIP ViT-L/14-336 & 32 & 26.60 & 77.57 & 21.33 & \underline{44.40} \\
        PLLaVA~\cite{pllava} & 13B & CLIP ViT-L/14-336 & 16 & 35.30 & 76.96 & 16.50 & 32.44\\
        VideoLLaMA2~\cite{videollama2} & 7B & CLIP ViT-L/14-336 & 16 & \underline{50.04} & \textbf{83.84} & 26.17 & \textbf{65.12} \\
        VILA1.5~\cite{vila} & 13B & SigLIPViT-SO-14-384 & 8 & \textbf{58.46} & \underline{81.88} & \textbf{63.33} & 35.03 \\
        \midrule
        Gemini-1.5-Pro~\cite{team2024gemini} & - & - & 16 & 75.27 & 79.25 & 83.83 & 63.96 \\
        GPT-4o~\cite{openai2024gpt4ocard} & - & - & 16 & 81.15 & 90.95 & 82.00 & 71.58 \\
        \midrule
        Human & - & - & - & 95.14 & 93.29 & 90.17 & 87.43 \\
        \bottomrule
    \end{tabular}}
    \caption{Performance comparison of existing MLLMs on \vidhalluc, covering action hallucination (ACH), temporal sequence hallucination (TSH), and scene transition hallucination (STH) tasks. For the STH task, we assign a weight of 0.6 to the classification task and 0.4 to the description task. The numbers in the table represent accuracy percentages (\%). \textbf{Bold} numbers denote the best performance, and \underline{underlined} numbers indicate the second-best performance.}    
    \label{tab:model_comparison}
    \vspace{-1em}
\end{table*}

\subsection{Alignment and Upsampling of Saliency Maps}
To align the DINOv2 saliency map with the original visual features, we reshape $S_{\text{DINOv2}}$ into a 2D grid corresponding to the spatial dimensions of the DINOv2 tokens:
\begin{equation} 
S_{\text{DINOv2}}^{\text{grid}} = \text{Reshape}(S_{\text{DINOv2}}, H_{\text{DINOv2}}, W_{\text{DINOv2}}),
\end{equation}
where $H_{\text{DINOv2}}$ and $W_{\text{DINOv2}}$ are the height and width of the DINOv2 token grid. We then upsample the saliency map to match the spatial dimensions of the visual features from the original visual encoder:
\begin{equation} 
S_{\text{DINOv2}}^{\text{upsampled}} = \text{Interpolate}(S_{\text{DINOv2}}^{\text{grid}}, (H_{\text{visual}}, W_{\text{visual}})),
\end{equation}
where $H_{\text{visual}}$ and $W_{\text{visual}}$ denote the height and width of the visual feature grid, corresponding to the spatial dimensions of the features extracted by the original visual encoder.
We then flatten the upsampled saliency map back into a vector and normalize it with the sigmoid function:
\begin{equation} 
S_{\text{DINOv2}}^{\text{normalized}} = \sigma (\text{Flatten}(S_{\text{DINOv2}}^{\text{upsampled}}) \in \mathbb{R}^{L_{\text{visual}}}).
\end{equation}

\subsection{Reweighting Visual Features}


Finally, we enhance the visual features by reweighting them with the saliency map:

\begin{equation} 
F_{\text{visual}}^{\text{reweighted}} = F_{\text{visual}} \odot S_{\text{DINOv2}}^{\text{normalized}},
\end{equation}
where $\odot$ denotes element-wise multiplication. This adaptive reweighting strategy enables DINOv2 to enhance key visual features by directly focusing on areas highlighted by the saliency map, thereby mitigating hallucinations while preserving the original feature representation.

%% file: sec/5_experiments.tex
\section{Experiments}


We evaluate ten state-of-the-art MLLMs on \vidhalluc, including eight open-source and two proprietary models (Section~\ref{eval-vidhalluc}).\ Our results reveal that most MLLMs exhibit notable vulnerabilities on \vidhalluc. To further evaluate the quality of \vidhalluc, we also conduct a human evaluation of our benchmark. We recruit four participants, each answering a randomly assigned half of the ACH, TSH, and STH queries, ensuring that each query is covered by two individuals. The questions, answer formats, and evaluation metrics are identical to those used for the models to ensure fairness.\
We then assess the performance of our DINO-HEAL on \vidhalluc~with five different MLLMs, demonstrating its effectiveness in enhancing the robustness of these models against various types of hallucinations (Section~\ref{eval-dino-heal}).\ During inference, we preserve each model’s original configuration, including conversation mode, hyperparameters, and frame count.\ Following standard practices~\citep{videollama2, videollama3}, we set the temperature to 0, top\_k to 1, and disable sampling for all models to avoid randomness in response generation.
For proprietary models, we sample 16 frames.\ Implementation details and full results are in the Supplementary Material.

\begin{figure}
    \centering
    \includegraphics[width=0.49\textwidth]{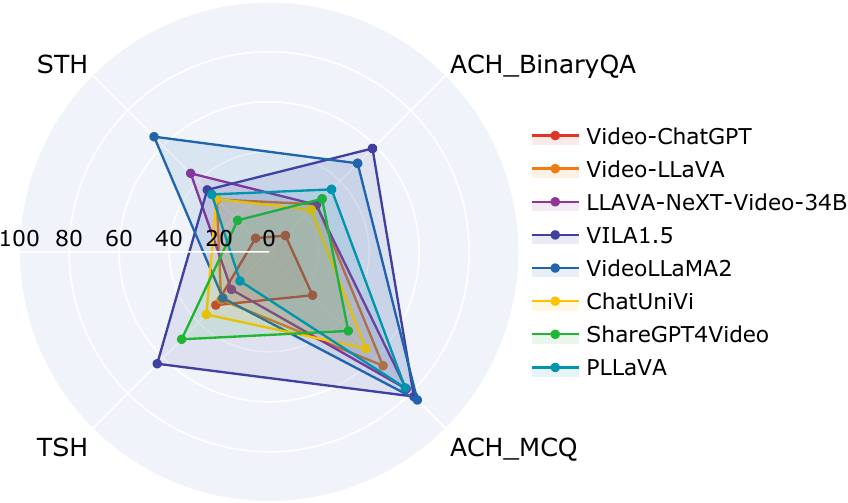}
    \caption{
    Comparative results on \vidhalluc~across various hallucination types.
    ACH, TSH, and STH indicate action, temporal sequence, and scene transition hallucination, respectively.
    }
    \label{fig:radar}
    \vspace{-1.5em}
\end{figure}

\subsection{Evaluation on \textbf{\vidhalluc}}
\label{eval-vidhalluc}

Table~\ref{tab:model_comparison} presents the performance of all tested MLLMs on \vidhalluc, showing both accuracy and score metrics as percentages, reflecting the occurrence of hallucinations across various tasks.\ For the ACH task, we observe that most models score at least 20\% higher on MCQs compared to binary QA, even though both question types are based on the same video and the same ground truth action.\ This significant difference is due to the MCQ design, which includes the ground truth action, an adversarial (semantically similar) action, and two distractors (irrelevant actions).
In contrast, binary QA presents either the ground truth or an adversarial action. Due to high semantic similarity, the model struggles to answer ``No'' to adversarial actions, as it lacks the comparative context of MCQs and must rely solely on action semantics. Higher MCQ accuracy suggests that when multiple options are available, the model can leverage additional contextual cues to differentiate semantically similar actions, reducing hallucination risks. Without this context in binary QA, the model is more prone to confusion and errors.

For TSH, most models score below 50\%, revealing significant challenges in distinguishing between semantically similar but distinct actions over time, particularly when these actions appear similar to the visual encoder.
For example, models often struggle to differentiate between successive actions, such as ``lifting a cup" followed by ``placing a cup down," when they occur in rapid sequence within the same video.
Notably, in nearly 50\% incorrect cases, models perceive only a single action throughout the entire video, failing to detect multiple actions or transitions.
Similarly, for STH, scores are generally low, with most models failing to surpass 40\%, highlighting substantial difficulties in accurately detecting and segmenting scene changes.

We also observe that model size and the number of input frames do not correlate directly with performance. For instance, LLaVA-NeXT-Video-34B, which processes 32 frames, performs significantly worse on each task compared to VILA1.5-13B, which operates with only 8 frames. Additionally, models with higher-resolution visual encoders generally outperform those with lower resolutions.\ Specifically, models using the CLIP ViT-L/14-336 encoder achieve higher scores across nearly all metrics compared to those using the lower-resolution CLIP ViT-L/14-224 encoder.

Proprietary models generally outperform open-source models across most tasks, with GPT-4o standing out in particular. However, GPT-4o still falls short of human performance, showing a 13.99\% gap in ACH binary QA and a 15.85\% gap in STH. In contrast, Gemini-1.5-Pro does not consistently surpass the best open-source models. For instance, its STH score (63.96\%) is slightly lower than that of VideoLLaMA2 (65.12\%). These results highlight the need for improvement even in top-performing proprietary models. 

Because GPT-4o plays a key role in the MCQ component by generating plausible distractors, we investigate whether it gains an unfair advantage by recognizing its own generated options. To assess this, we test GPT-4o on 300 videos with corresponding MCQs. Among these, GPT-4o correctly identifies its own generated distractors only twice, misidentifies them 33 times, and responds with ``I don’t know'' in 265 instances. These results suggest that GPT-4o is unlikely to gain an unfair advantage during evaluation.

\subsection{Evaluation of DINO-HEAL on \textbf{\vidhalluc}}
\label{eval-dino-heal}
To evaluate the effectiveness of DINO-HEAL in mitigating hallucinations, we use Video-LLaVA~\cite{videollava}, Video-ChatGPT~\cite{videochatgpt}, VILA~\cite{vila}, and VideoLLaMA2~\cite{videollama2} as the foundation backbones. 
\begin{table}[ht]
    \centering
    \resizebox{0.48\textwidth}{!}{
    \begin{tabular}{lcccc}
        \toprule
         \multirow{2.5}{*}{\makebox[0.15\linewidth][c]{\textbf{Models}}} & \multicolumn{2}{c}{\textbf{Acc. on ACH}} & \multirow{2.5}{*}{\textbf{TSH}} &\multirow{2.5}{*}{\textbf{STH}} \\
        \cmidrule(r){2-3}
         & \textbf{Binary QA} & \textbf{MCQ} \\
        \midrule
        Video-ChatGPT & 9.50 & 24.58 & 30.17 & 7.70 \\
        \rowcolor{gray!20}
        \scriptsize +DINO-HEAL & 13.96\textsubscript{+4.46} & 28.81\textsubscript{+4.23} & 41.83\textsubscript{+11.66} & 8.20\textsubscript{+0.5} \\
        Video-LLaVA  & 26.84 & 64.45 & 27.17 & 29.60 \\
        \rowcolor{gray!20}
        \scriptsize +DINO-HEAL & 33.80\textsubscript{+6.96} & 66.25\textsubscript{+1.8} & 28.50\textsubscript{+1.33} & 31.42\textsubscript{+1.82} \\
        ShareGPT4Video  & 29.96 & 44.78 & 49.50 & 17.83 \\
        \rowcolor{gray!20}
        \scriptsize +DINO-HEAL & 30.41\textsubscript{+0.45} & 44.43\textsubscript{-0.35} & 55.33\textsubscript{+5.83} & 18.39\textsubscript{+0.56} \\
        VILA1.5 & 58.46 & 81.88 & 63.33 &  35.03 \\
        \rowcolor{gray!20}
        \scriptsize +DINO-HEAL & \textbf{60.63}\textsubscript{+2.17} & 81.85\textsubscript{-0.03} & \textbf{64.17}\textsubscript{+0.84} & 36.15\textsubscript{+1.12} \\
        VideoLLaMA2 & 50.04 & 83.84 & 26.17 & 65.12 \\
        \rowcolor{gray!20}
        \scriptsize +DINO-HEAL & 50.01\textsubscript{-0.03} & \textbf{83.84}\textsubscript{+0.0} & 44.50\textsubscript{+18.33} & \textbf{66.17}\textsubscript{+1.05} \\
        \bottomrule
    \end{tabular}}
    \caption{Performance comparison of models on action hallucination (ACH), temporal sequence hallucination (TSH), and scene transition hallucination (STH) tasks, with and without DINO-HEAL. Improvements from DINO-HEAL are shown as subscripts. \textbf{Bold} numbers denote the best performance after applying DINO-HEAL.}
    \label{tab:mitigated_comparison}
    \vspace{-1em}
\end{table}
Table~\ref{tab:mitigated_comparison} shows that DINO-HEAL enables substantial improvements across various hallucination types, achieving an average gain of 3.02\%.

In the ACH task, DINO-HEAL leads to notable accuracy gains, with Video-LLaVA and Video-ChatGPT showing increases of +6.96\% and +4.46\% in Binary QA, respectively.\ 
For the TSH task, DINO-HEAL significantly enhances temporal coherence, with Video-ChatGPT achieving an +11.66\% gain and VideoLLaMA2 a substantial +18.33\% increase.\ 
In contrast, improvements in the STH task are more limited, with an average 1.01\% increase in performance.\
We attribute the phenomenon to the inherent focus of DINOv2, the visual encoder, on foreground objects rather than background elements or scene transitions, as it is primarily trained to extract visual features through discriminative self-supervised learning.\
This foreground bias enhanced DINO-HEAL's impact on action recognition, driving significant gains in ACH and TSH. However, it limits its effectiveness for background scene understanding, as models may overlook background changes crucial for accurate scene boundary detection.

DINO-HEAL is also designed to be compatible with a variety of visual encoders beyond CLIP, such as LanguageBind-Video~\cite{languagebind} in Video-LLaVA and SigLIP~\cite{siglip} in VILA1.5.\ After applying DINO-HEAL, both Video-LLaVA and VILA1.5 demonstrate consistent improvements across nearly all tasks, highlighting DINO-HEAL's versatility and effectiveness across diverse model architectures.

%% file: sec/6_conclusion.tex
\section{Conclusion and Future Work}

We introduce \vidhalluc, the largest benchmark for evaluating action, temporal sequence, and scene transition hallucinations in MLLMs for video understanding.\ We also present DINO-HEAL, a novel training-free method to mitigate MLLM hallucinations by enhancing the visual encoder's focus on salient spatial regions during inference, improving model robustness against hallucinations without additional training. Extensive experiments show the effectiveness of DINO-HEAL in mitigating hallucinations across models. Future work includes expanding hallucination categories to assess models in diverse settings and enhancing DINO-HEAL with a dual-stream design integrating both spatial and temporal saliency for improved video understanding.
\clearpage

%% file: sec/X_suppl.tex
\clearpage
\renewcommand{\thesection}{\Alph{section}}
\setcounter{section}{0}
\setcounter{page}{1}

\maketitlesupplementary


\section{Additional Related Works}
Different neural network architectures, trained on distinct datasets, develop unique inductive biases that influence their feature representation capabilities. For instance, the CLIP series~\citep{clip, evaclip, siglip}, pretrained with text-image contrastive alignment, excels at capturing global semantic features. In contrast, the DINO series~\cite{dinov2, dino}, pretrained with vision-only contrastive learning, specializes in fine-grained perception and object-level details.
Prior research shows that integrating features from complementary networks can lead to a more balanced and robust model behavior~\citep{geigle2023doesfitallcomplementarity, rodriguezopazo2023unveilingbackboneeffectsclip}. 

Building on these strengths, COMM~\cite{jiang2023clip} demonstrates the effectiveness of integrating features from different layers of CLIP and DINOv2 to enhance the visual capabilities of multimodal large language models (MLLMs).
By leveraging the complementary nature of these architectures, COMM achieves improved performance across various visual tasks. Similarly, CLIP-DINOiser~\cite{wysoczanska2024clip} uses localization priors from DINO to refine CLIP’s global image features, resulting in smoother and more precise outputs for semantic segmentation tasks.
Furthermore, Nguyen \textit{et al.}~\cite{nguyen2024exploring} explore the multi-level features of DINO to fine-tune the final block of CLIP.
This approach not only tackles the challenge posed by the limited scale of training datasets in deepfake detection but also enhances model interpretability through the use of attention mechanisms. These studies collectively highlight the potential of combining the strengths of diverse architectures to overcome individual limitations, achieve superior performance, and improve explainability in complex visual tasks.
However, these methods often rely on specific architectures and require additional adjustments or training for integration, which limits their flexibility and scalability.

To address these limitations, our proposed DINO-HEAL offers a flexible and architecture-agnostic solution that can be applied at the inference stage without modifying the underlying model structures.
Unlike previous approaches, DINO-HEAL does not require additional training parameters or fine-tuning, making it particularly suitable for mitigating hallucinations in resource-constrained scenarios.

\section{Implementation Details}

\subsection{Data Collection on Existing Datasets}

To apply our data collection pipeline to the selected datasets, we employ a structured process for segmenting and pairing videos. Specifically, for ActivityNet~\cite{activitynet} and YouCook2~\cite{youcook2}, videos are divided into multiple action-based segments. We then compute similarities between all segment pairs within each video to identify pairs meeting the similarity criteria. For VALOR32K~\cite{valor}, videos are randomly paired, and their similarities are calculated to determine if they satisfy the conditions. The resulting filtered video pairs, characterized by high semantic similarity but low visual similarity, form the core of our dataset, enabling effective investigation into hallucination phenomena.

\subsection{Prompts for Different Hallucination Tasks}

The following subsections show the prompts we use to test models on each hallucination task: action hallucination (ACH), temporal sequence hallucination (TSH), and scene transitional hallucination (STH).

\subsubsection{Prompt for Binary QA in ACH}

\small{
\begin{tcolorbox}[colback=gray!20!white, colframe=black, boxrule=0.5mm, arc=2mm, breakable]
\begin{verbatim}
<Video>
Is the primary action in the video 
{Action}?
Only answer with "No" or "Yes".
\end{verbatim}
\end{tcolorbox}}
\normalsize

\noindent The placeholder \{Action\} in the prompt is dynamically replaced with a specific action, such as `turning the steering wheel'. To further enhance the diversity of the benchmark, each Binary QA question is randomly assigned one of four templates: \textit{``Is the prominent action in the video \{Action\}?"}, \textit{``Does the video primarily feature \{Action\}?"}, \textit{``Is the key action shown in the video \{Action\}?"}, or \textit{``Is the primary action in the video \{Action\}?"} These variations introduce linguistic diversity while preserving the semantic meaning.

\subsubsection{Prompt for MCQs in ACH}


\small
\begin{tcolorbox}[colback=gray!20!white, colframe=black, boxrule=0.5mm, arc=2mm, breakable]    
\begin{verbatim}
<Video>
"Question": "What is the prominent 
action in the video?" Please select
the correct answer (one or more options),
only return the choice letter (i.e., A, 
B, C, D) of your answer(s).

"Choices":
"A": "{Action A}"
"B": "{Action B}"
"C": "{Action C}"
"D": "{Action D}"
\end{verbatim}
\end{tcolorbox}
\normalsize

\noindent The placeholders \{Action A\}, \{Action B\}, \{Action C\}, and \{Action D\} are dynamically replaced with specific actions, such as ``wakeboarding,” ``changing gears,” ``adjusting the rearview mirror,” and ``turning the steering wheel.” To introduce linguistic diversity and enhance the robustness of the benchmark, each MCQ is randomly assigned one of the following templates: \textit{``What is the prominent action in the video?", ``What is the key action shown in the video?"}, \textit{``What is the primary action in the video?"}, or \textit{``What is the predominant action captured in the video?"}. These variations ensure that the benchmark reflects a range of natural question formulations while maintaining consistency in meaning.

\subsubsection{Prompt for Sorting Questions in TSH}

\small
\begin{tcolorbox}[colback=gray!20!white, colframe=black, boxrule=0.5mm, arc=2mm, breakable]    
\begin{verbatim}
<Video>
Below are two actions in the video:
Action A. {Action A}
Action B. {Action B}

Sort these two actions in the order they 
occur in the video, and return which 
action happens before which one. For 
example, "Action A before Action B" or 
"Action B before Action A". If you only 
detect one action of these two in the 
video, return that action.
\end{verbatim}
\end{tcolorbox}
\normalsize
\noindent The placeholders \{Action A\}, \{Action B\} are replaced with specific actions. For instance, with the actions of skiing and driving a car, the prompt will look as the following example:

\vspace{0.1cm}
\noindent \textit{``Below are two actions in the video: Action A. skiing, Action B. driving a car. Sort these two actions in the order they 
occur in the video, and return which action happens before which one. For example, `Action A before Action B' or `Action B before Action A'. If you only detect one action of these two in the video, return that action.''}

\subsubsection{Prompt for Open-ended Questions in STH}
\small
\begin{tcolorbox}[colback=gray!20!white, colframe=black, boxrule=0.5mm, arc=2mm, breakable]    
\begin{verbatim}
<Video>
A scene change is defined as a 
significant transition in the overall 
environment or location within the 
video. This means a change from one 
distinct setting to another, such as 
moving from a kitchen to a living room, 
or from indoors to outdoors. Watch the 
given video and determine if a scene 
change occurs. If there is a scene 
change, respond in the format: "Scene 
change: Yes, Locations: from [location] 
to [location2]." If no change occurs, 
respond: "Scene change: No, Locations: 
None".
\end{verbatim}
\end{tcolorbox}
\normalsize

\subsection{Additional Dataset Statistics}
\begin{figure}[h]
    \centering
    \includegraphics[width=0.5\textwidth]{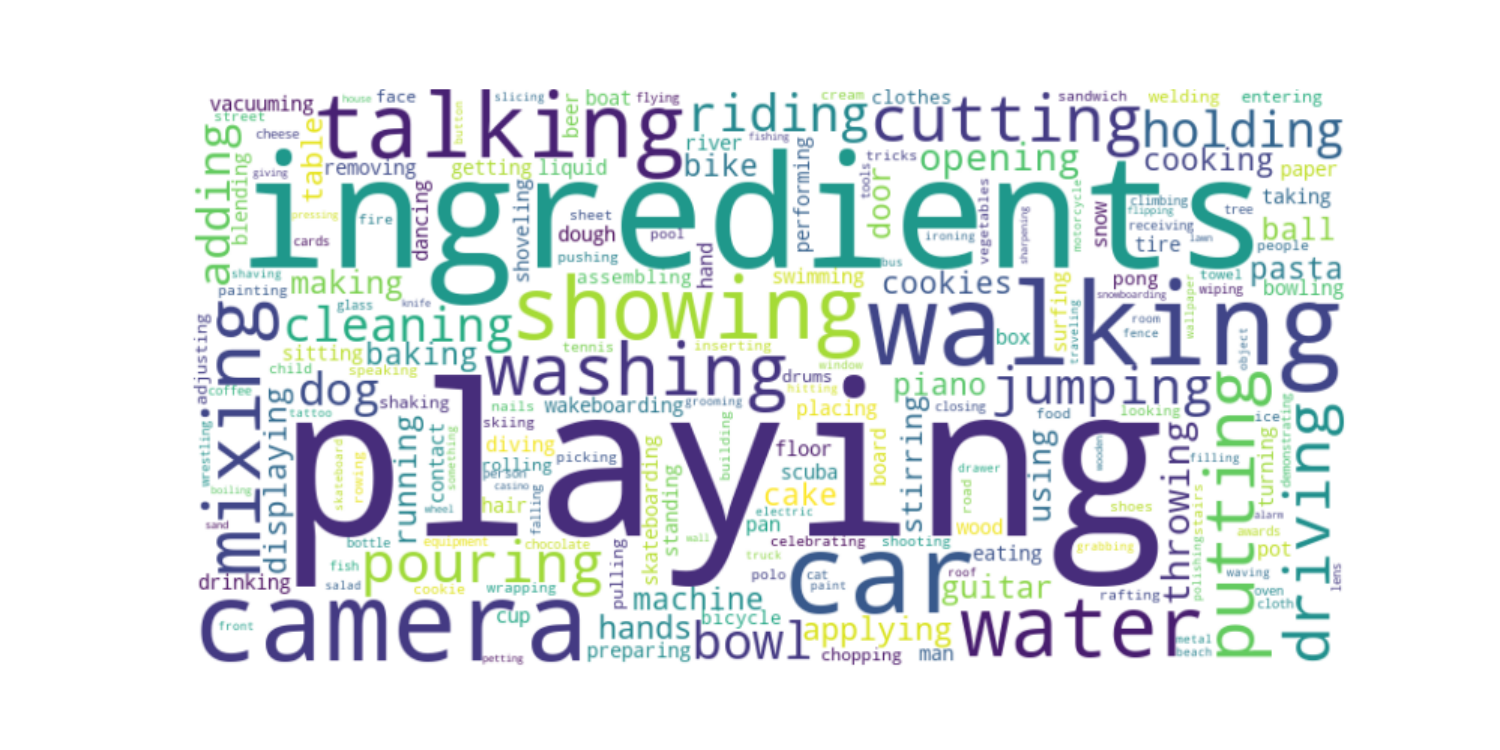}
    \vspace{-1cm}
    \caption{The wordcloud of \vidhalluc.}
    \label{fig:wordcloud}
\end{figure}

Figure~\ref{fig:wordcloud} displays a word cloud of our benchmark, providing a more intuitive presentation of \vidhalluc.
As shown in the figure, the questions in our benchmark are diverse and prominently feature action-related terms, such as ``playing,'' ``walking,'' ``cutting,'' ``cleaning,'' and ``jumping.'' These terms highlight the dynamic nature of the video content in our benchmark, emphasizing the focus on activities and interactions within videos. The word cloud reflects the breadth of actions and events covered, including activities like ``mixing ingredients,'' ``pouring water,'' ``riding,'' and ``driving.'' This diversity aligns with our goal of capturing the challenges associated with action recognition, temporal coherence, and scene understanding in videos. 
We believe our benchmark can effectively reveal potential hallucination issues in MLLMs, especially those related to understanding complex actions in dynamic video content.

\section{More Quantitative Results on \vidhalluc}
We provide additional metrics and detailed scores of state-of-the-art MLLM performance on \vidhalluc.\
For open source models, we include Video-ChatGPT~\cite{videochatgpt}, Video-LLaVA~\cite{videollava}, ShareGPT4Video~\cite{sharegpt4video}, Chat-UniVi~\cite{chatunivi}, LLaVA-NeXT-Video~\cite{llavanextvideo}, PLLaVA~\cite{pllava}, VideoLLaMA2~\cite{videollama2}, and VILA 1.5~\cite{vila}. For proprietary models, we select Gemini-1.5-Pro~\cite{team2024gemini}, and GPT-4o~\cite{achiam2023gpt}.

\paragraph{Quantitative results on ACH.} 
Table~\ref{tab:ach_halluc} presents two distinct versions of the quantitative results for these models on the ACH task. The binary QA and MCQ scores are computed by dividing the number of correct answers by the total number of questions, following the metric defined as:

\begin{equation}
    \text{Accuracy} = \frac{N_\text{correct}}{N_\text{total}},
\end{equation}
where $N_\text{correct}$ and $N_\text{total}$ denote non-negative counts of correct answers and total answers.
In contrast, binary QA Pair and MCQ Pair scores are based on a stricter criterion requiring both questions in a pair to be answered correctly.
This stricter evaluation ensures that the model fully understands both semantically similar but visually different videos.

As mentioned in the main paper, an interesting observation is that the accuracy for MCQ is higher than that for binary questions.
This result defies common intuition, as one might expect binary questions, with only two possible answers, to be inherently simpler and, therefore, yield higher accuracy compared to MCQs, which involve selecting from multiple options.
This discrepancy suggests that models may leverage contextual or comparative cues more effectively in MCQ scenarios, while binary questions might require more precise reasoning or direct understanding, exposing potential weaknesses in model comprehension.

\begin{table}[t]
    \centering
    \resizebox{0.48\textwidth}{!}{
    \begin{tabular}{lcccc}
        \toprule
         \textbf{Models} & \textbf{Binary QA} & \textbf{Binary QA Pair} & \textbf{MCQ} & \textbf{MCQ Pair} \\
        \midrule
        Video-ChatGPT~\cite{videochatgpt} & 9.50 & 0.19 & 24.58 & 5.56 \\
        Video-LLaVA~\cite{videollava}  & 26.84 & 9.87 & 64.45 & 40.34 \\
        ShareGPT4Video~\cite{sharegpt4video}  & 29.96 & 10.65 & 44.78 & 19.18 \\
        Chat-UniVi~\cite{chatunivi} & 23.77 & 6.39 & 54.79 & 29.37\\
        LLaVA-NeXT-Video~\cite{llavanextvideo} & 26.60 & 12.00 & 77.57 & 60.03 \\
        PLLaVA~\cite{pllava} & 35.30 & 16.26 & 76.96 & 59.94 \\
        VideoLLaMA2~\cite{videollama2} & 50.04 & 29.09 & \underline{83.84} & \underline{69.85} \\
        VILA1.5~\cite{vila} & 58.46 & 37.77 & 81.88 & 67.95 \\
        \midrule
        Gemini-1.5-Pro~\cite{team2024gemini} & \underline{75.27} & \underline{59.10} & 79.25 & 63.36 \\
        GPT-4o~\cite{achiam2023gpt} & \textbf{81.15} & \textbf{66.79} & \textbf{90.95} & \textbf{83.00} \\
        \bottomrule
    \end{tabular}}
    \caption{Performance comparison of existing models on action hallucination (ACH). The numbers in the table represent accuracy percentages (\%). \textbf{Bold} numbers denote the best performance, and \underline{underlined} numbers indicate the second-best performance.}
    \label{tab:ach_halluc}
\end{table}

\paragraph{Quantitative results on STH}
In STH, we benchmark MLLMs with the new criterion that evaluates both the classification of the scene and whether the model describes it in the correct sequence.
For the classification scores, we use the Matthews correlation coefficient (MCC) to evaluate the model predictions against the ground truth labels.

\begin{equation}
   \frac{n_{11} \times n_{10} - n_{01} \times n_{00}}{\sqrt{(n_{11} + n_{01}) (n_{11} + n_{00})(n_{10} + n_{01})(n_{10} + n_{00})}},
    \label{eq:mcc} 
\end{equation}

\noindent where $A \in \{0\ (\text{False}),\ 1\ (\text{True})\}$ represents the actual condition, $P \in \{0\ (\text{Negative}),\ 1\ (\text{Positive})\}$ represents the predicted condition, and $n_{AP}$ denotes non-negative counts.
To adjust MCC to range between 0 and 1 and to further penalize models that consistently answer only ``Yes'' or only ``No'', we apply the transformation in order to adjust MCC to range between 0 and 1, obtaining the classification score $\text{Score}_{\text{cls}}$:
\begin{equation}
    \text{Score}_\text{cls} = \left( \frac{\text{MCC} + 1}{2} \right)^2.
    \label{eq:cls_score}
\end{equation}

\noindent The description task measures the model's ability to accurately identify and articulate the information of the scene.
To evaluate this, scene descriptions are extracted from both the model’s output and the ground truth, structured for direct comparison.
We then calculate the cosine similarity $S$ between the SimCSE~\cite{simcse} embeddings of the corresponding scenes. Based on this similarity measure, each scene description score is calculated as:

\begin{equation}
    \text{Score}_{\text{desc}} = 
    \begin{cases} 
        \hfill 0, & \text{if } S \leq \text{THR}_{\text{low}} \\
        \frac{\sigma(S) - \sigma(\text{THR}_{\text{low}})}{\sigma(1) - \sigma(\text{THR}_{\text{low}})}, & \text{if } S > \text{THR}_{\text{low}}
    \end{cases},
    \label{eq:desc_score}
\end{equation}

\begin{figure*}[h]
    \centering
    \includegraphics[width=0.9\textwidth]{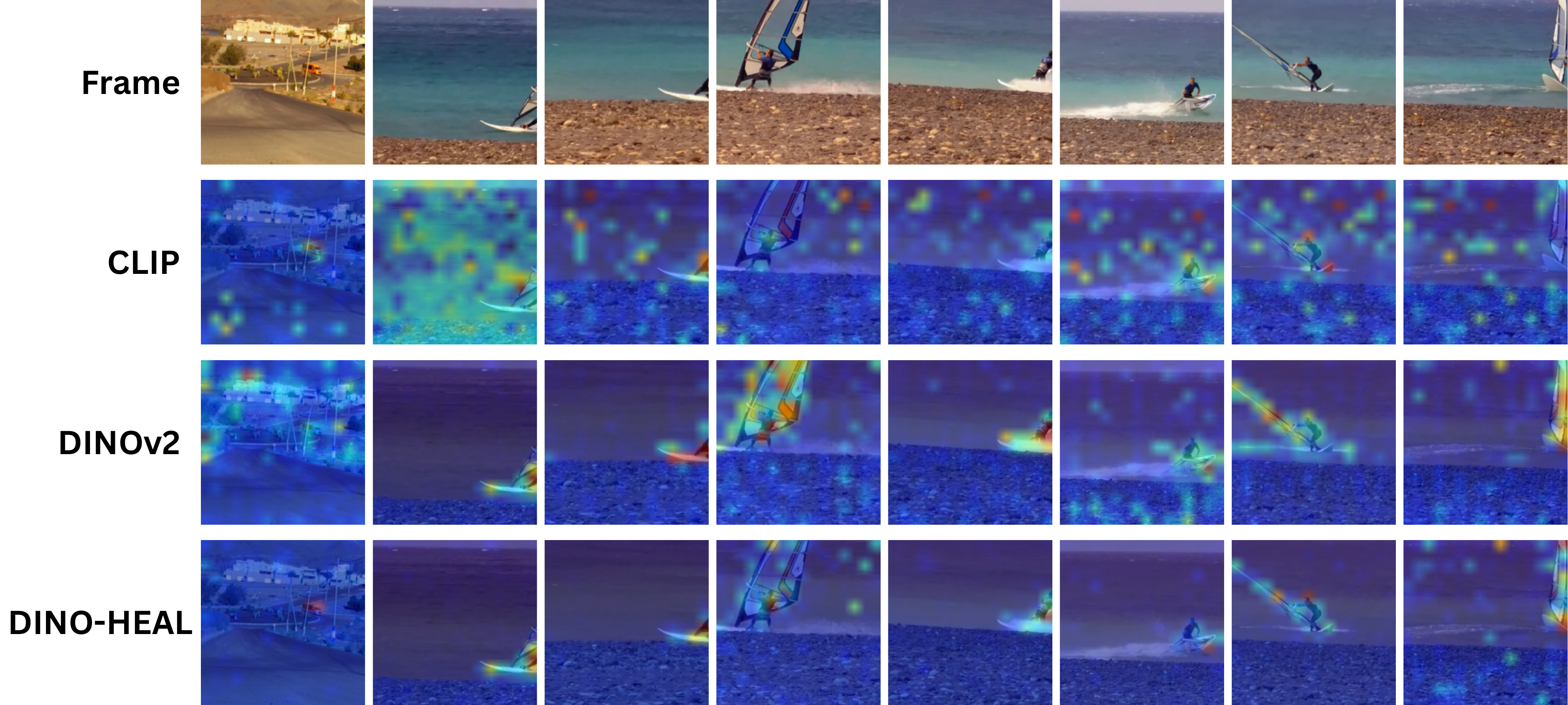}
    \caption{The visualization of saliency maps.}
    \label{fig:saliency-maps}
\end{figure*}

\noindent where $S$ denotes the cosine similarity between the SimCSE~\cite{simcse} embeddings of the corresponding scenes, \( \text{THR}_{\text{low}} \) represents the minimum threshold for assigning a score, and $\sigma$ is the Sigmoid function.
The overall description score is calculated as the average of the score for the ``from'' and ``to'' scenes.
Finally, the overall evaluation score is computed as a weighted sum of the classification score and the normalized description score:

\begin{equation}
    \text{Score}_{\text{overall}} = \alpha \times \text{Score}_{\text{cls}} + (1 - \alpha) \times \text{Score}_{\text{desc}}.
    \label{eq:overall_score}
\end{equation}

\begin{table}[t]
    \centering
    \resizebox{0.48\textwidth}{!}{
    \begin{tabular}{lcccc}
        \toprule
         \textbf{Models} & \textbf{$\text{Score}_\text{cls}$} & \textbf{$\text{Score}_\text{desc}$} & \textbf{$\text{Score}_\text{overall}$} \\
        \midrule
        Video-ChatGPT~\cite{videochatgpt} & 5.07 & 11.65 & 7.70 \\
        Video-LLaVA~\cite{videollava}  & 25.00 & 36.50 & 29.60 \\
        ShareGPT4Video~\cite{sharegpt4video}  & 29.55 & 0.26 & 17.83 \\
        Chat-UniVi~\cite{chatunivi} & 30.12 & 29.50 & 29.87 \\
        LLaVA-NeXT-Video~\cite{llavanextvideo} & 55.91 & 27.14 & 44.40 \\
        PLLaVA~\cite{pllava} & 29.86 & 36.32 & 32.44 \\
        VideoLLaMA2~\cite{videollama2} & \textbf{87.43} & 31.67 & \underline{65.12} \\
        VILA1.5~\cite{vila} & 25.00 & 50.07 & 35.03 \\
        \midrule
        Gemini-1.5-Pro~\cite{team2024gemini} & 71.88 & \underline{52.08} & 63.96 \\
        GPT-4o~\cite{achiam2023gpt} & \underline{80.17} & \textbf{58.69} & \textbf{71.58} \\
        \bottomrule
    \end{tabular}}
    \caption{Performance comparison of existing models on scene transition hallucination (STH). We assign a weight of 0.6 to the classification task and 0.4 to the description task. The numbers in the table represent accuracy percentages (\%). \textbf{Bold} numbers denote the best performance, and \underline{underlined} numbers indicate the second-best performance.}
    \label{tab:sth_halluc}
\end{table}

\noindent Table~\ref{tab:sth_halluc} summarizes the decomposed scores of classification and description scores for the STH category.
An interesting observation is that both Video-LLaVA and VILA 1.5 achieve a classification score of 25\% by consistently answering ``Yes'' to all questions, irrespective of their actual ability to recognize transitions between locations.
This pattern, highlighted in the 10th row of Table~\ref{tab:tsh-example}, exposes a critical limitation in both the MCC metric and the model themselves.
Their reliance on default affirmative response reveals a superficial understanding of spatial transition and suggests a lack of deeper reasoning. \label{sec:tsh-detail}

Future work should focus on developing mechanisms to penalize such oversights and promote consistency in model behavior, ensuring that metrics better reflect genuine understanding and performance. This includes designing metrics or loss functions that discourage uniform responses and promote adaptive reasoning based on context.

\section{DINO-HEAL}

\subsection{Saliency Analysis}
Figure~\ref{fig:saliency-maps} illustrates the input frames, saliency maps generated by CLIP and DINOv2, and the adjusted saliency map produced by DINO-HEAL.
The saliency maps generated by CLIP often show significant noise, which can be attributed to its inductive bias toward capturing global contextual frame information.
This characteristic, while beneficial for understanding broader scene-level features, can result in a lack of focus on specific, spatially important regions within the frame.
In contrast, DINOv2 demonstrates a stronger capacity to localize and emphasize key objects within the scene, leveraging its vision-only contrastive learning to identify fine-grained details efficiently.
The adjusted saliency maps created by DINO-HEAL reflect an integration of these two approaches, balancing the strengths of both CLIP and DINOv2.
By mitigating the noisiness of CLIP's global focus and incorporating the precise localization capabilities of DINOv2, DINO-HEAL effectively emphasizes spatially significant features.
This result strongly supports our hypothesis that DINO-HEAL serves as a complementary mechanism to CLIP, enhancing its ability to prioritize critical regions and improving overall spatial feature representation.

\subsection{Additional Results on \vidhalluc}
Tables~\ref{tab:ach_mitigated} and \ref{tab:sth_mitigated} show the results of the ACH and STH tasks when baseline models are augmented with our hallucination mitigation method, DINO-HEAL.
A particularly noteworthy observation is the significant improvement in scores for Video-LLaVA and VILA 1.5 on the STH task.
Previously, these models consistently defaulted to answering ``Yes" regardless of the correct location-based response.
With the integration of DINO-HEAL, however, they demonstrate an improved ability to discern and appropriately respond with "No" when necessary, as elaborated in  Section~\ref{sec:tsh-detail}.
This indicates a meaningful enhancement in their spatial reasoning and decision-making capabilities.
This improvement underscores the potential of DINO-HEAL to refine spatial reasoning and address their shortcomings effectively.

Tables~\ref{tab:dinoheal-ach-example} and~\ref{tab:dinoheal-tsh-example} further detail the results for Video-ChatGPT, Video-LLaVA, and VideoLLaMA2 on the ACH and TSH tasks, comparing performance with and without DINO-HEAL.
For the ACH task, we use the binary QA accuracy metric.
Without DINO-HEAL, none of the models correctly predict all question pairs, though VideoLLaMA2 is able to infer the second question pair accurately.
When DINO-HEAL is applied, however, all models can predict both pairs accurately, showcasing the method’s effectiveness in mitigating hallucinations. On the TSH task, we observe further improvements.
Initially, Video-ChatGPT recognizes that two actions occur simultaneously, while the ground truth sequence is ``starting a fire'' followed by ``gutting a fish''.
Video-LLaVA and VideoLLaMA2 only identify one action, ``starting a fire''.
After integrating DINO-HEAL, all three models correctly identify and sequence both actions, underscoring the method's ability to enhance temporal understanding in complex tasks.

\begin{table}[t]
    \centering
    \resizebox{0.48\textwidth}{!}{
    \begin{tabular}{lcccc}
        \toprule
         \textbf{Models} & \textbf{Binary QA} & \textbf{Binary QA Pair} & \textbf{MCQ} & \textbf{MCQ Pair} \\
        \midrule
        Video-ChatGPT & 9.50 & 0.19 & 24.58 & 5.56 \\
        \rowcolor{gray!20}
        \scriptsize +DINO-HEAL & 13.96\textsubscript{+4.46} & 0.42\textsubscript{+0.23} & 28.81\textsubscript{+4.23} & 6.76\textsubscript{+1.20} \\
        Video-LLaVA  & 26.84 & 9.87 & 64.45 & 40.34 \\
        \rowcolor{gray!20}
        \scriptsize +DINO-HEAL & 33.80\textsubscript{+6.96} & 14.17\textsubscript{+4.3} & 66.25\textsubscript{+1.8} & 41.64\textsubscript{+1.3} \\
        ShareGPT4Video  & 29.96 & 10.65 & 44.78 & 19.18 \\
        \rowcolor{gray!20}
        \scriptsize +DINO-HEAL & 30.41\textsubscript{+0.45} & 9.73\textsubscript{-0.92} & 44.43\textsubscript{-0.35} & 18.62\textsubscript{-0.56} \\
        VILA1.5 & 58.46 & 37.77 & 81.88 & 67.95 \\
        \rowcolor{gray!20}
        \scriptsize +DINO-HEAL & \textbf{60.63}\textsubscript{+2.17} & \textbf{40.34}\textsubscript{+2.57} & 81.85\textsubscript{-0.03} & 67.90\textsubscript{-0.05} \\
        VideoLLaMA2 & 50.04 & 29.09 & 83.84 & 69.85 \\
        \rowcolor{gray!20}
        \scriptsize +DINO-HEAL & 50.01\textsubscript{-0.03} & 29.07\textsubscript{-0.02} & \textbf{83.84}\textsubscript{+0.0} & \textbf{69.85}\textsubscript{+0.0} \\
        \bottomrule
    \end{tabular}}
    \caption{Performance comparison of models on action hallucination (ACH), with and without DINO-HEAL. Improvements from DINO-HEAL are shown as subscripts. \textbf{Bold} numbers denote the best performance after applying DINO-HEAL.}
    \label{tab:ach_mitigated}
\end{table}

\begin{table}[ht]
    \centering
    \resizebox{0.48\textwidth}{!}{
    \begin{tabular}{lccc}
        \toprule
         \textbf{Models} & \textbf{$\text{Score}_\text{cls}$} & \textbf{$\text{Score}_\text{desc}$} & \textbf{$\text{Score}_\text{overall}$} \\
        \midrule
        Video-ChatGPT & 5.07 & 11.65 & 7.70 \\
        \rowcolor{gray!20}
        \scriptsize +DINO-HEAL & 5.56\textsubscript{+0.49} & 12.15\textsubscript{+0.5} & 8.20\textsubscript{+0.5} \\
        Video-LLaVA  & 25.00 & 36.50 & 29.60 \\
        \rowcolor{gray!20}
        \scriptsize +DINO-HEAL & 27.89\textsubscript{+2.89} & 35.18\textsubscript{-2.61} & 30.81\textsubscript{+0.69} \\
        ShareGPT4Video  & 29.55 & 0.26 & 17.83 \\
        \rowcolor{gray!20}
        \scriptsize +DINO-HEAL & 28.80\textsubscript{-0.75} & 2.78\textsubscript{+2.52} & 18.39\textsubscript{+0.56} \\
        VILA1.5 & 25.00 & 50.07 & 35.03 \\
        \rowcolor{gray!20}
        \scriptsize +DINO-HEAL & 26.66\textsubscript{+1.66} &\textbf{50.38}\textsubscript{+0.21} & 36.15\textsubscript{+1.12} \\
        VideoLLaMA2 & 87.43 & 31.67 & 65.12 \\
        \rowcolor{gray!20}
        \scriptsize +DINO-HEAL & \textbf{89.19}\textsubscript{+1.76} & 31.63\textsubscript{-0.04} & \textbf{66.17}\textsubscript{+1.05} \\
        \bottomrule
    \end{tabular}}
    \caption{ Performance comparison of models on scene transition hallucination (STH), with and without DINO-HEAL. We assign a weight of 0.6 to the classification task and 0.4 to the description task. Improvements from DINO-HEAL are shown as subscripts. \textbf{Bold} numbers denote the best performance  applying DINO-HEAL.}
    \label{tab:sth_mitigated}
\end{table}

\begin{table*}[h]
    \resizebox{\textwidth}{!}{
    \centering
    \begin{tabular}{p{0.12\textwidth}p{0.8\textwidth}}
    \toprule
    & \textbf{DINO-HEAL Example, Action Hallucination}\\
    \midrule
        \parbox[t]{0.12\textwidth}{\centering \textbf{Clip 1}} 
        & \parbox[c]{0.8\textwidth}{\includegraphics[width=0.8\textwidth]{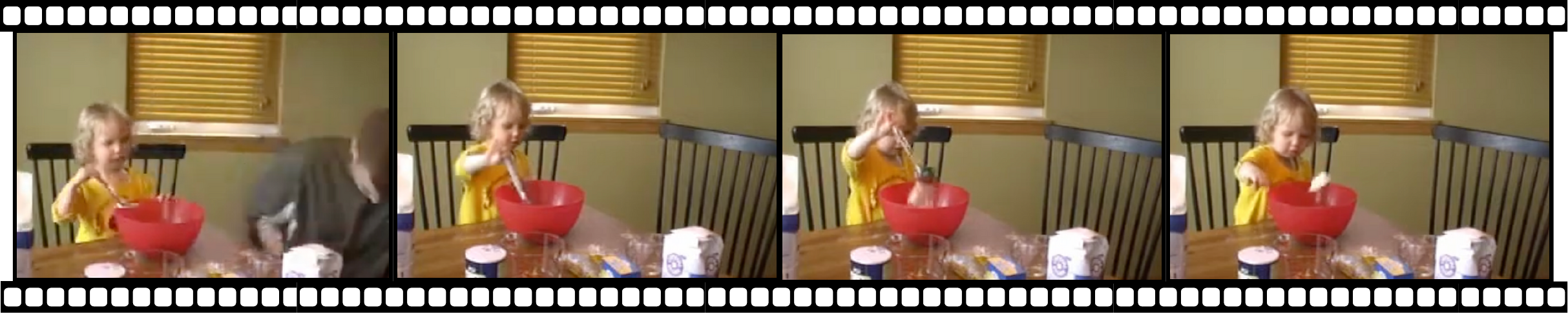}} \\
        \parbox[t]{0.12\textwidth}{\centering \textbf{Clip 2}} 
        & \parbox[c]{0.8\textwidth}{\includegraphics[width=0.8\textwidth]{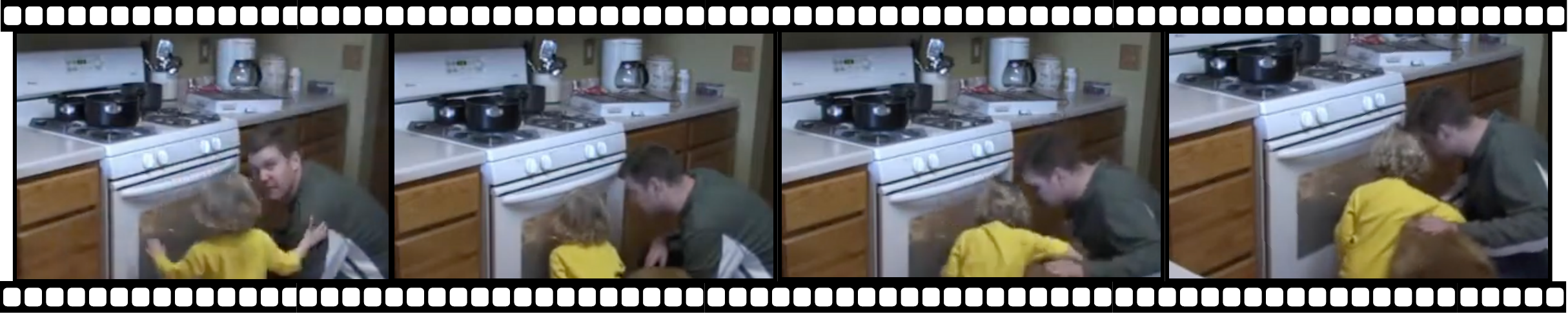}
        \vspace{0.15em}
        } \\
        Question 1 & Is the primary action in the video mixing the ingredients? \\
        Question 2 & Is the primary action in the video watching the oven? \\
        \midrule
        Ground & \textit{Answer for Question 1 with Clip 1:} \textcolor{OliveGreen}{{\textbf{Yes}}}.\\
        Truth & \textit{Answer for Question 1 with Clip 2:} \textcolor{OliveGreen}{{\textbf{No}}}. \\
        & \textit{Answer for Question 2 with Clip 1:} \textcolor{OliveGreen}{{\textbf{No}}}. \\
        & \textit{Answer for Question 2 with Clip 2:} \textcolor{OliveGreen}{{\textbf{Yes}}}. \\
        \midrule
        Video-ChatGPT & \textit{Answer for Question 1 with Clip 1:}  \textcolor{Red}{{\textbf{Yes}}}, the primary action in the video is mixing the ingredients. \newline \textit{Answer for Question 1 with Clip 2:} \textcolor{Red}{{\textbf{Yes}}}, the primary action in the video is mixing the ingredients.\\
        & \textit{Answer for Question 2 with Clip 1:} \textcolor{Red}{{\textbf{No}}}. The man and the child are making cookies in the kitchen. \\
        & \textit{Answer for Question 2 with Clip 2:} \textcolor{Red}{{\textbf{No}}}. \\
        \midrule
        Video-ChatGPT + \small \textbf{DINO-HEAL} & \textit{Answer for Question 1 with Clip 1}: \textcolor{OliveGreen}{{\textbf{Yes}}}.\newline \textit{Answer for Question 1 with Clip 2:} \textcolor{OliveGreen}{{\textbf{No}}}. \newline \textit{Answer for Question 2 with Clip 1:} \textcolor{OliveGreen}{{\textbf{No}}}. \newline \textit{Answer for Question 2 with Clip 2:} \textcolor{OliveGreen}{{\textbf{Yes}}}. \\
        \midrule
        Video-LLaVA & \textit{Answer for Question 1 with Clip 1:} \textbf{\textcolor{Red}{Yes}}.\newline \textit{Answer for Question 1 with Clip 2:} \textbf{\textcolor{Red}{Yes}}. \newline \textit{Answer for Question 2 with Clip 1:} \textbf{\textcolor{Red}{Yes}}. \newline \textit{Answer for Question 2 with Clip 2:} \textbf{\textcolor{Red}{Yes}}. \\
        \midrule
        Video-LLaVA + \small \textbf{DINO-HEAL} & \textit{Answer for Question 1 with Clip 1}: \textcolor{OliveGreen}{{\textbf{Yes}}}.\newline \textit{Answer for Question 1 with Clip 2:} \textcolor{OliveGreen}{{\textbf{No}}}. \newline \textit{Answer for Question 2 with Clip 1:} \textcolor{OliveGreen}{{\textbf{No}}}. \newline \textit{Answer for Question 2 with Clip 2:} \textcolor{OliveGreen}{{\textbf{Yes}}}. \\
        \midrule
        VideoLLaMA2 & \textit{Answer for Question 1 with Clip 1:} \textbf{\textcolor{Red}{Yes}}.\newline \textit{Answer for Question 1 with Clip 2:} \textbf{\textcolor{Red}{Yes}}. \newline \textit{Answer for Question 2 with Clip 1:} \textbf{\textcolor{OliveGreen}{No}}. \newline \textit{Answer for Question 2 with Clip 2:} \textbf{\textcolor{OliveGreen}{Yes}}. \\
        \midrule
        VideoLLaMA2 + \small \textbf{DINO-HEAL} & \textit{Answer for Question 1 with Clip 1}: \textcolor{OliveGreen}{{\textbf{Yes}}}.\newline \textit{Answer for Question 1 with Clip 2:} \textcolor{OliveGreen}{{\textbf{No}}}. \newline \textit{Answer for Question 2 with Clip 1:} \textcolor{OliveGreen}{{\textbf{No}}}. \newline \textit{Answer for Question 2 with Clip 2:} \textcolor{OliveGreen}{{\textbf{Yes}}}. \\
        \bottomrule
        
    \end{tabular}}
    \caption{An example of the action hallucination category of \vidhalluc~dataset with and without DINO-HEAL.
    \textcolor{OliveGreen}{\textbf{Green}} text indicates correct answers, and \textcolor{red}{\textbf{red}} text indicates incorrect answers.}
    \label{tab:dinoheal-ach-example}
\end{table*}

\begin{table*}[h]
    \resizebox{\textwidth}{!}{
    \centering
    \begin{tabular}{p{0.14\textwidth}p{0.8\textwidth}}
    \toprule
    & \textbf{DINO-HEAL Example, Temporal Sequence Hallucination}\\
    \midrule  
        & \parbox[c]{0.8\textwidth}{\includegraphics[width=0.8\textwidth]{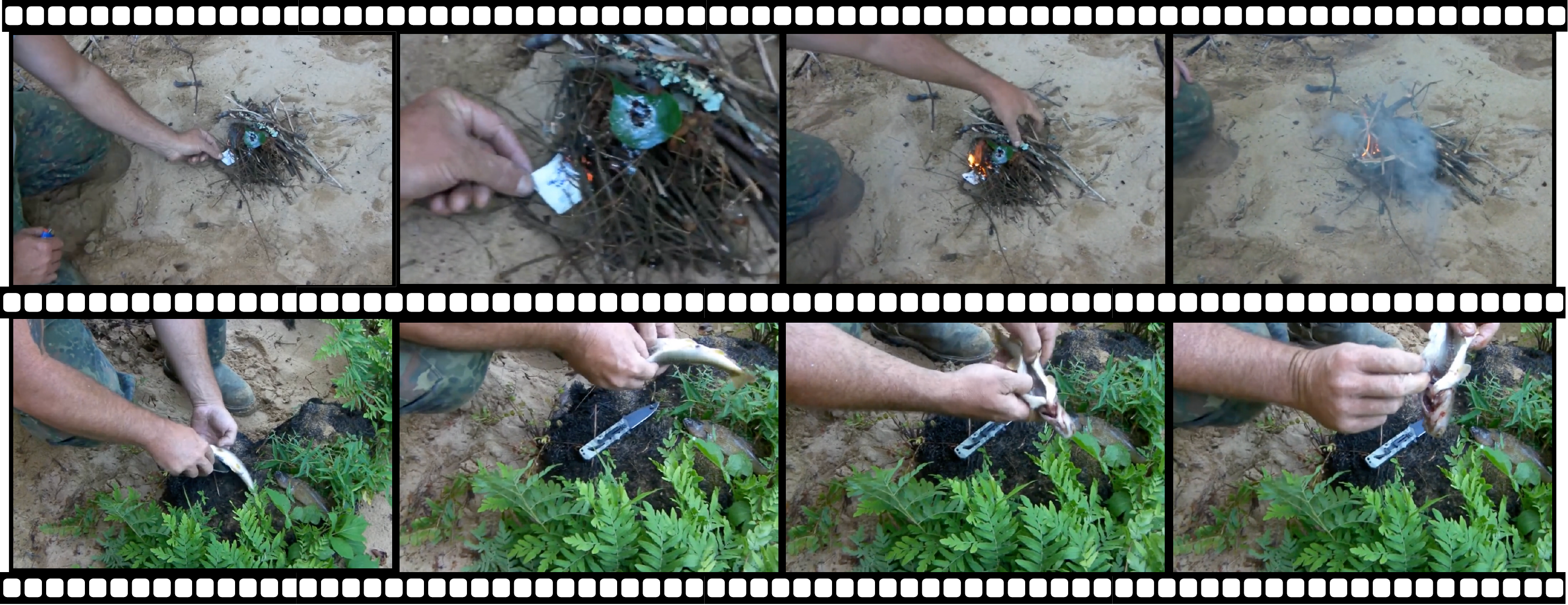}
        \vspace{0.15em}
        } \\
        Question & Below are two actions in the video: \\
        & \textbf{Action A. gutting a fish}\\
        & \textbf{Action B. starting a fire} \\
        & Sort these two actions in the order they occur in the video, and return which action happen before which one. For example, `Action A before Action B' or `Action B before Action A'. If you only detect one action of these two in the video, return that action. \\
        \midrule
        Ground Truth & \textcolor{OliveGreen}{{\textbf{BA}}}.\\
        \midrule
        Video-ChatGPT & In the video, the man is seen gutting a fish and starting a fire. The order of these two actions is not clear, as they both \textbf{\textcolor{Red}{occur simultaneously}}. \\
        \midrule
        Video-ChatGPT + \small \textbf{DINO-HEAL} & In the video, the man \textbf{\textcolor{OliveGreen}{starts a fire}} by striking a match on a piece of wood. \textbf{After that}, he goes to the beach and starts \textcolor{OliveGreen}{\textbf{gutting a fish}}.\\
        \midrule
        Video-LLaVA & Action \textbf{\textcolor{red}{B}}. starting a fire. \\
        \midrule
        Video-LLaVA \newline + \small \textbf{DINO-HEAL} & Action \textbf{\textcolor{OliveGreen}{B}} happens \textbf{before} Action \textbf{\textcolor{OliveGreen}{A}}.\\
        \midrule
        VideoLLaMA2 & I only detect one action in the video, which is Action \textbf{\textcolor{Red}{B}}: starting a fire. \\
        \midrule
        VideoLLaMA2 \newline + \small \textbf{DINO-HEAL} & Action \textbf{\textcolor{OliveGreen}{B}}. starting a fire happens \textbf{before} Action \textbf{\textcolor{OliveGreen}{A}}. gutting a fish. \\
        \midrule
        ShareGPT4Video & Answer: 'Action \textbf{\textcolor{OliveGreen}{B}} happens \textbf{before} Action \textbf{\textcolor{OliveGreen}{A}}' \\
        \midrule
        ShareGPT4Video \newline + \small \textbf{DINO-HEAL} & Answer: 'Action \textbf{\textcolor{OliveGreen}{B}} happens \textbf{before} Action \textbf{\textcolor{OliveGreen}{A}}' \\
        \midrule
        VILA1.5 & Action \textbf{\textcolor{OliveGreen}{B}}. starting a fire happens before Action \textbf{\textcolor{OliveGreen}{A}}. gutting a fish. \\
        \midrule
        VILA1.5 \newline + \small \textbf{DINO-HEAL} & Action \textbf{\textcolor{OliveGreen}{B}}. starting a fire happens before Action \textbf{\textcolor{OliveGreen}{A}}. gutting a fish. \\
        \bottomrule
        
    \end{tabular}}
    \caption{An example from the temporal sequence hallucination category of \vidhalluc~dataset with and without DINO-HEAL.
    \textcolor{OliveGreen}{\textbf{Green}} text indicates correct answers, and \textcolor{red}{\textbf{red}} text indicates incorrect answers.}
    \label{tab:dinoheal-tsh-example}
\end{table*}

\begin{table*}[h]
    \resizebox{\textwidth}{!}{
    \centering
    \begin{tabular}{p{0.12\textwidth}p{0.8\textwidth}}
    \toprule
    & \textbf{\vidhalluc~Example, Action Hallucination}\\
    \midrule
        \parbox[t]{0.12\textwidth}{\centering \textbf{Clip 1}} 
        & \parbox[c]{0.8\textwidth}{\includegraphics[width=0.8\textwidth]{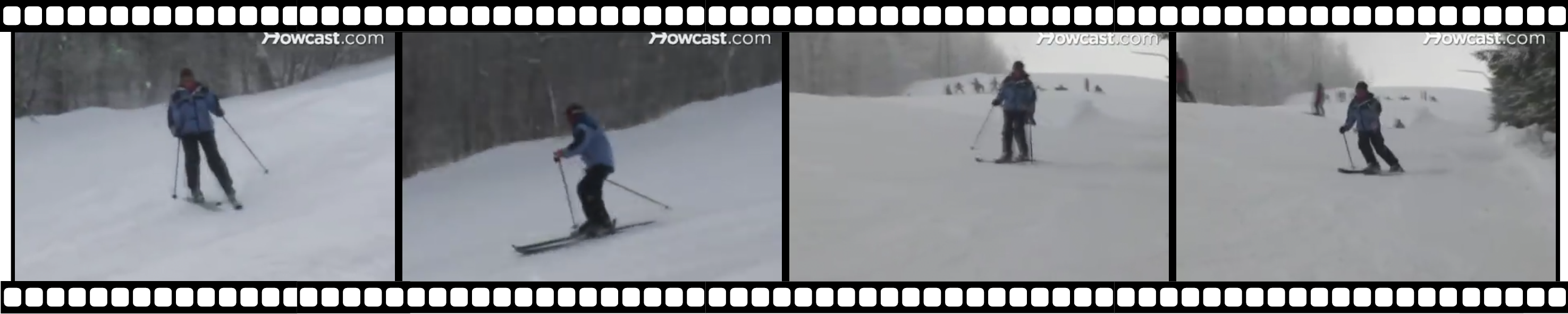}} \\
        \parbox[t]{0.12\textwidth}{\centering \textbf{Clip 2}} 
        & \parbox[c]{0.8\textwidth}{\includegraphics[width=0.8\textwidth]{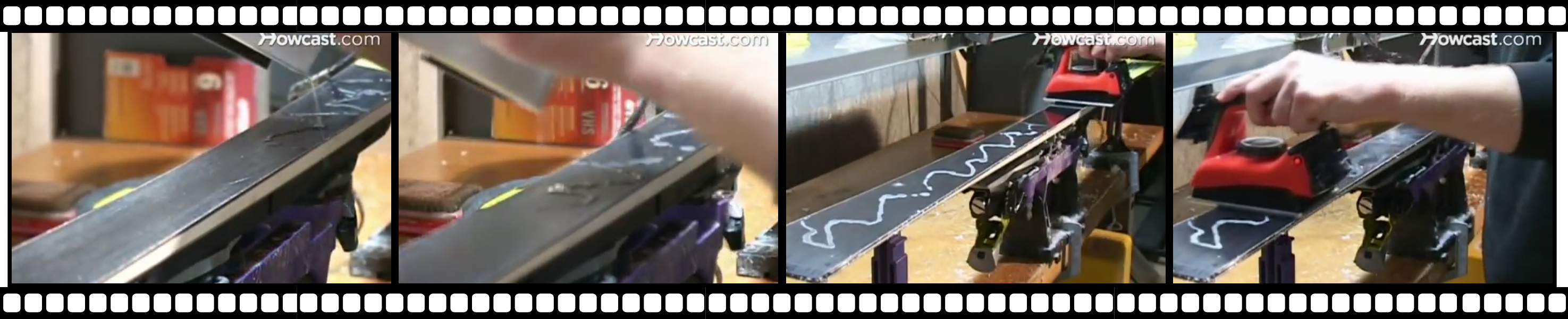}
        \vspace{0.15em}
        } \\
        Question 1 & Is ironing wax on the ski the main activity in the video? \\
        Question 2 & Is the primary action in the video skiing? \\
        \midrule
        Ground & \textit{Answer for Question 1 with Clip 1:} \textcolor{OliveGreen}{{\textbf{No}}}.\\
        Truth & \textit{Answer for Question 1 with Clip 2:} \textcolor{OliveGreen}{{\textbf{Yes}}}. \\
        & \textit{Answer for Question 2 with Clip 1:} \textcolor{OliveGreen}{{\textbf{Yes}}}. \\
        & \textit{Answer for Question 2 with Clip 2:} \textcolor{OliveGreen}{{\textbf{No}}}. \\
        \midrule
        Chat-Univi~\cite{chatunivi} & \textit{Answer for Question 1 with Clip 1}: \textcolor{Red}{{\textbf{No}}}, the main activity in the video is skiing down a snowy slope.\newline \textit{Answer for Question 1 with Clip 2:} \textcolor{Red}{{\textbf{No}}}, the main activity in the video is not ironing wax on the ski. The video showcases a person using a variety of tools and techniques to create snowboards. \newline \textit{Answer for Question 2 with Clip 1:} \textcolor{Red}{{\textbf{Yes}}}, the primary action in the video is skiing. \newline \textit{Answer for Question 2 with Clip 2:} \textcolor{Red}{{\textbf{Yes}}}, the primary action in the video is skiing. \\
        \midrule
        PLLaVA~\cite{pllava} & \textit{Answer for Question 1 with Clip 1:} \textbf{\textcolor{Red}{Yes}}.\newline \textit{Answer for Question 1 with Clip 2:} \textbf{\textcolor{Red}{Yes}}. \newline \textit{Answer for Question 2 with Clip 1:} \textbf{\textcolor{Red}{Yes}}. \newline \textit{Answer for Question 2 with Clip 2:} \textbf{\textcolor{Red}{Yes}}. \\
        \midrule
        Video- & \textit{Answer for Question 1 with Clip 1:}  \textcolor{Red}{{\textbf{Yes}}}, the main activity in the video is ironing wax on the ski.\\
        ChatGPT~\cite{videochatgpt} & \textit{Answer for Question 1 with Clip 2:} \textcolor{Red}{{\textbf{No}}}, the main activity in the video is not ironing wax on the ski. The video shows a man using a machine to scrape wood. \\
        & \textit{Answer for Question 2 with Clip 1:} \textcolor{Red}{{\textbf{No}}}, the primary action in the video is not skiing. \\
        & \textit{Answer for Question 2 with Clip 2:} \textcolor{Red}{{\textbf{No}}}, the primary action in the video is not skiing. The video shows a man working on a piece of wood with a tool. \\
        \midrule
        LLaVA-NeXT-Video~\cite{llavanextvideo} & \textit{Answer for Question 1 with Clip 1:}  \textcolor{Red}{{\textbf{Yes}}}. \newline \textit{Answer for Question 1 with Clip 2:} \textcolor{Red}{{\textbf{Yes}}}. \newline \textit{Answer for Question 2 with Clip 1:} \textcolor{Red}{{\textbf{Yes}}}. \newline \textit{Answer for Question 2 with Clip 2:} \textcolor{Red}{{\textbf{Yes}}}. \\
        \bottomrule
        
    \end{tabular}}
    \caption{An example from the action hallucination category of \vidhalluc~dataset.
    \textcolor{OliveGreen}{\textbf{Green}} text indicates correct answers, and \textcolor{red}{\textbf{red}} text indicates incorrect answers.}
    \label{tab:ach-example}
\end{table*}

\begin{table*}[h]
    \resizebox{\textwidth}{!}{
    \centering
    \begin{tabular}{p{0.17\textwidth}p{0.8\textwidth}}
    \toprule
    & \textbf{\vidhalluc~Example, Temporal Sequence Hallucination}\\
    \midrule  
        & \parbox[c]{0.8\textwidth}{\includegraphics[width=0.8\textwidth]{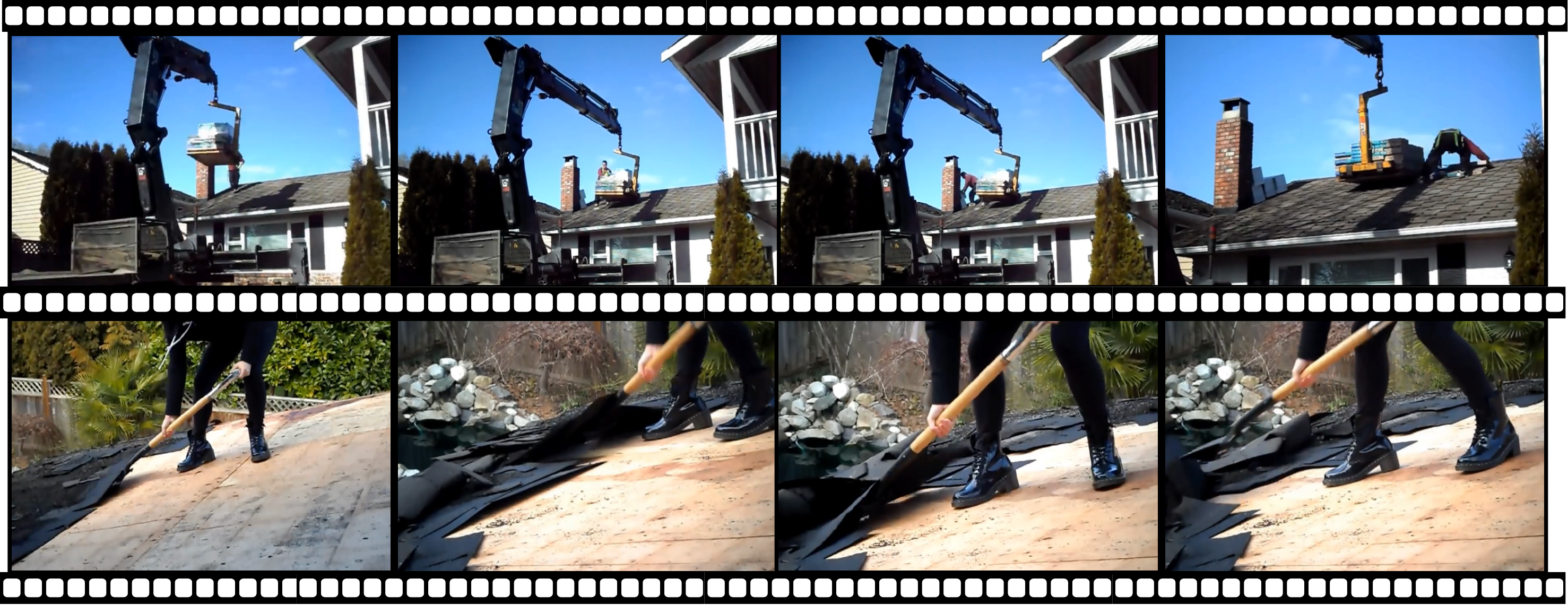}
        \vspace{0.15em}
        } \\
        Question & Below are two actions in the video: \\
        & \textbf{Action A. unloading shingles onto the roof}\\
        & \textbf{Action B. removing old shingles} \\
        & Sort these two actions in the order they occur in the video, and return which action happens before which one. For example, `Action A before Action B' or `Action B before Action A'. If you detect only one of the two actions in the video, return that action. \\
        \midrule
        Ground Truth & \textcolor{OliveGreen}{{\textbf{AB}}}.\\
        \midrule
        LLaVA-NeXT-Video~\cite{llavanextvideo} & Action \textbf{\textcolor{red}{B}}. removing old shingles.\\
        \midrule
        Chat-UniVi~\cite{chatunivi} & Based on the video, Action \textbf{\textcolor{red}{B}} (removing old shingles) occurs \textbf{before} Action \textbf{\textcolor{red}{A}} (unloading shingles onto the roof).\\
        \midrule
        Video-LLaVA~\cite{videollava} & Action \textbf{\textcolor{red}{B}}. removing old shingles happens \textbf{before} Action \textbf{\textcolor{red}{A}}. unloading shingles onto the roof. \\
        \midrule
        PLLaVA~\cite{pllava} & Action \textbf{\textcolor{red}{B}}. removing old shingles.\\
        \midrule
        Gemini-1.5-Pro~\cite{team2024gemini} & The order of actions is \textbf{\textcolor{red}{BA}}:\newline 1. **B. removing old shingles:** A person is shown using a shovel to remove old shingles from the roof.\newline 2. **A. unloading shingles onto the roof:** A crane is used to lift new shingles onto the roof.\\
        \midrule
        GPT-4o~\cite{achiam2023gpt} & \textbf{\textcolor{red}{BA}}.\\
        \midrule
        Video-ChatGPT~\cite{videochatgpt} & In the video, the person is first seen unloading shingles onto the roof, and then they are seen removing old shingles. The order in which these actions occur is \textbf{\textcolor{OliveGreen}{unloading shingles onto the roof}} first, \textbf{followed by} \textbf{\textcolor{OliveGreen}{removing old shingles}}. \\
        \midrule
        ShareGPT4Video~\cite{sharegpt4video} & Answer: `Action \textbf{\textcolor{OliveGreen}{A}} happens \textbf{before} Action \textbf{\textcolor{OliveGreen}{B}}'.\\
        \midrule
        VideoLLaMA2~\cite{videollama2} & Action \textbf{\textcolor{OliveGreen}{A}}. unloading shingles onto the roof happens \textbf{before} Action \textbf{\textcolor{OliveGreen}{B}}. removing old shingles. \\
        \midrule
        VILA1.5~\cite{vila} & Action \textbf{\textcolor{red}{B}}. removing old shingles happens \textbf{before} Action \textbf{\textcolor{red}{A}}. unloading shingles onto the roof. \\
        \bottomrule
        
    \end{tabular}}
    \caption{An example from the temporal sequence hallucination category of the \vidhalluc~dataset.
    \textcolor{OliveGreen}{\textbf{Green}} text indicates correct answers, and \textcolor{red}{\textbf{red}} text indicates incorrect answers.}
    \label{tab:tsh-example}
\end{table*}

\begin{table*}[h]
    \resizebox{\textwidth}{!}{
    \centering
    \begin{tabular}{p{0.17\textwidth}p{0.8\textwidth}}
    \toprule
    & \textbf{\vidhalluc~Example, Scene Transition Hallucination}\\
    \midrule  
        & \parbox[c]{0.8\textwidth}{\includegraphics[width=0.8\textwidth]{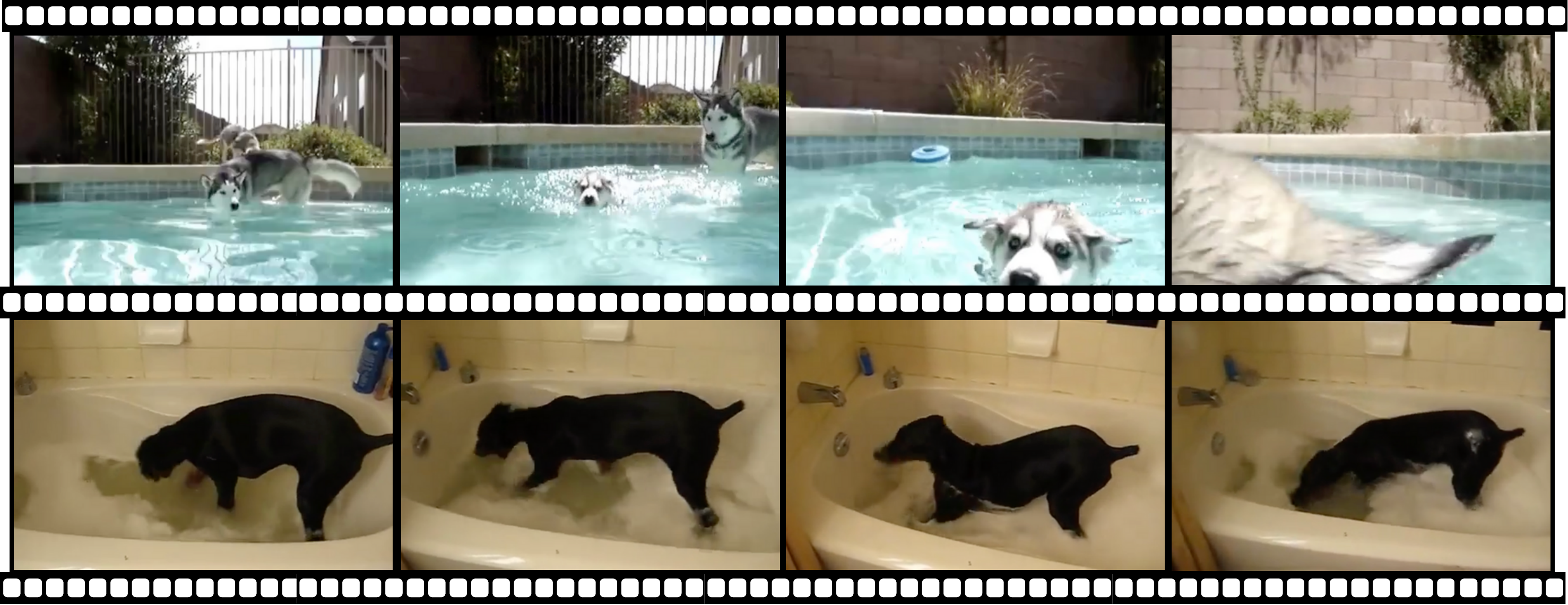}
        \vspace{0.15em}
        } \\
        Question & A scene change is defined as a significant transition in the overall environment or location within the video. This means a change from one distinct setting to another, such as moving from a kitchen to a living room or from indoors to outdoors. Watch the given video and determine if a scene change occurs. If there is a scene change, respond in the format: `Scene change: Yes, Locations: from [location1] to [location2].' If no change occurs, respond: `Scene change: No, Locations: None'. \\
        \midrule
        Ground Truth & Scene change: \textcolor{OliveGreen}{{\textbf{Yes}}}; Locations: from \textbf{in a swimming pool} to \textbf{in a bathtub}.\\
        \midrule
        Video-ChatGPT~\cite{videochatgpt} & Scene change: \textcolor{Red}{{\textbf{No}}}; Locations: \textbf{None}. \\
         & \textit{$\text{Score}_\text{desc1}=0$, $\text{Score}_\text{desc2}=0$}\\
        \midrule
        LLaVA-NeXT-Video~\cite{llavanextvideo} & Scene change: \textcolor{OliveGreen}{{\textbf{Yes}}}; Locations: from \textbf{pool} to \textbf{bathtub}. \newline \textit{$\text{Score}_\text{desc1}=0.727$, $\text{Score}_\text{desc2}=0.892$}\\
        \midrule
        Chat-UniVi~\cite{chatunivi} & Scene change: \textcolor{OliveGreen}{{\textbf{Yes}}}; Locations: from \textbf{indoor pool} to \textbf{outdoor pool}.\\
        & \textit{$\text{Score}_\text{desc1}=0.514$, $\text{Score}_\text{desc2}=0$}\\
        \midrule
        Video-LLaVA~\cite{videollava} & Scene change: \textcolor{OliveGreen}{{\textbf{Yes}}}; Locations: from \textbf{the pool} to \textbf{the bathtub}. \\
        & \textit{$\text{Score}_\text{desc1}=0.862$, $\text{Score}_\text{desc2}=0.961$} \\
        \midrule
        PLLaVA~\cite{pllava} & Scene change: \textcolor{OliveGreen}{{\textbf{Yes}}}; Locations: from \textbf{indoor bathroom} to \textbf{indoor bathroom}.\\
        & \textit{$\text{Score}_\text{desc1}=0$, $\text{Score}_\text{desc2}=0.266$} \\
        \midrule
        ShareGPT4Video~\cite{sharegpt4video} & Scene change: \textcolor{Red}{{\textbf{No}}}; Locations: \textbf{None}.\\
         & \textit{$\text{Score}_\text{desc1}=0$, $\text{Score}_\text{desc2}=0$}\\
        \midrule
        VideoLLaMA2~\cite{videollama2} & Scene change: \textcolor{OliveGreen}{{\textbf{Yes}}}; Locations: from \textbf{pool} to \textbf{bathtub}. \\
         & \textit{$\text{Score}_\text{desc1}=0.727$, $\text{Score}_\text{desc2}=0.892$}\\
        \midrule
        VILA1.5~\cite{vila} & Scene change: \textcolor{OliveGreen}{{\textbf{Yes}}}; Locations: from \textbf{[location1]} to \textbf{[location2]}. \\
         & \textit{$\text{Score}_\text{desc1}=0$, $\text{Score}_\text{desc2}=0$}\\
        \midrule
        Gemini-1.5-Pro~\cite{team2024gemini} & Scene change: \textcolor{OliveGreen}{{\textbf{Yes}}}; Locations: from \textbf{pool} to \textbf{bathtub}. \\
         & \textit{$\text{Score}_\text{desc1}=0.727$, $\text{Score}_\text{desc2}=0.892$}\\
        \midrule
        GPT-4o~\cite{achiam2023gpt} & Scene change: \textcolor{OliveGreen}{{\textbf{Yes}}}; Locations: from \textbf{a swimming pool} to \textbf{a bathtub}.\\
         & \textit{$\text{Score}_\text{desc1}=0.941$, $\text{Score}_\text{desc2}=0.946$}\\
        \bottomrule
        
    \end{tabular}}
    \caption{An example from the scene transition hallucination category of the \vidhalluc~dataset.
    \textcolor{OliveGreen}{\textbf{Green}} text indicates correct answers, and \textcolor{red}{\textbf{red}} text indicates incorrect answers. Each model's description performance is evaluated using two scores: $\text{Score}_\text{desc1}$ and $\text{Score}_\text{desc2}$, derived from Equation~\ref{eq:desc_score}. These scores correspond to the model's ability to describe the two distinct scenes in the video accurately. The model's overall $\text{Score}_\text{desc2}$ is computed as the average of these two scores.}
    \label{tab:sth-example}
\end{table*}

\section{More Qualitative Results on \vidhalluc}
Tables~\ref{tab:ach-example},~\ref{tab:tsh-example}, and~\ref{tab:sth-example} present randomly selected examples showcasing multiple model responses for the ACH, TSH, and STH tasks, respectively.
For the ACH task, the Binary QA Pair metric is used, which applies a stricter evaluation criterion requiring both questions in a pair to be answered correctly.
In particular, the adversarially crafted pairs in the ACH category require MLLMs to discern between semantically similar actions, posing a significant challenge for accurate interpretation.
None of the models could predict both pairs accurately, highlighting the complexity of the ACH task. Moreover, Table~\ref{tab:tsh-example} details the description scores for each example.
These examples demonstrate that locations highly relevant to the ground truth achieve high scores.
For instance, the phrase ``in a swimming pool" exhibits strong semantic description similarity with ``pool" or ``a swimming pool," while ``in a bathtub" aligns well with ``the bathtub".
Conversely, locations with differing semantic descriptions or inaccurate interpretations receive lower description scores.
For example, ``in a swimming pool" has limited similarity to descriptions such as ``indoor bathroom" or ``indoor pool," and ``in a bathtub" similarly diverges from ``indoor bathroom".
These results highlight the importance of semantic alignment in achieving accurate description scores.

\clearpage


%% file: main.bbl
\begin{thebibliography}{77}
\providecommand{\natexlab}[1]{#1}
\providecommand{\url}[1]{\texttt{#1}}
\expandafter\ifx\csname urlstyle\endcsname\relax
  \providecommand{\doi}[1]{doi: #1}\else
  \providecommand{\doi}{doi: \begingroup \urlstyle{rm}\Url}\fi

\bibitem[Achiam et~al.(2023)Achiam, Adler, Agarwal, Ahmad, Akkaya, Aleman, Almeida, Altenschmidt, Altman, Anadkat, et~al.]{achiam2023gpt}
Josh Achiam, Steven Adler, Sandhini Agarwal, Lama Ahmad, Ilge Akkaya, Florencia~Leoni Aleman, Diogo Almeida, Janko Altenschmidt, Sam Altman, Shyamal Anadkat, et~al.
\newblock Gpt-4 technical report.
\newblock \emph{arXiv preprint arXiv 2303.08774}, 2023.

\bibitem[Alayrac et~al.(2022)Alayrac, Donahue, Luc, Miech, Barr, Hasson, Lenc, Mensch, Millican, Reynolds, Ring, Rutherford, Cabi, Han, Gong, Samangooei, Monteiro, Menick, Borgeaud, Brock, Nematzadeh, Sharifzadeh, Binkowski, Barreira, Vinyals, Zisserman, and Simonyan]{flamingo}
Jean-Baptiste Alayrac, Jeff Donahue, Pauline Luc, Antoine Miech, Iain Barr, Yana Hasson, Karel Lenc, Arthur Mensch, Katie Millican, Malcolm Reynolds, Roman Ring, Eliza Rutherford, Serkan Cabi, Tengda Han, Zhitao Gong, Sina Samangooei, Marianne Monteiro, Jacob Menick, Sebastian Borgeaud, Andrew Brock, Aida Nematzadeh, Sahand Sharifzadeh, Mikolaj Binkowski, Ricardo Barreira, Oriol Vinyals, Andrew Zisserman, and Karen Simonyan.
\newblock Flamingo: a visual language model for few-shot learning.
\newblock In \emph{NeurIPS}, 2022.

\bibitem[Brown et~al.(2020)Brown, Mann, Ryder, Subbiah, Kaplan, Dhariwal, Neelakantan, Shyam, Sastry, Askell, Agarwal, Herbert-Voss, Krueger, Henighan, Child, Ramesh, Ziegler, Wu, Winter, Hesse, Chen, Sigler, Litwin, Gray, Chess, Clark, Berner, McCandlish, Radford, Sutskever, and Amodei]{gpt3}
Tom Brown, Benjamin Mann, Nick Ryder, Melanie Subbiah, Jared~D Kaplan, Prafulla Dhariwal, Arvind Neelakantan, Pranav Shyam, Girish Sastry, Amanda Askell, Sandhini Agarwal, Ariel Herbert-Voss, Gretchen Krueger, Tom Henighan, Rewon Child, Aditya Ramesh, Daniel Ziegler, Jeffrey Wu, Clemens Winter, Chris Hesse, Mark Chen, Eric Sigler, Mateusz Litwin, Scott Gray, Benjamin Chess, Jack Clark, Christopher Berner, Sam McCandlish, Alec Radford, Ilya Sutskever, and Dario Amodei.
\newblock Language models are few-shot learners.
\newblock In \emph{NeurIPS}, 2020.

\bibitem[Chandrasegaran et~al.(2024)Chandrasegaran, Gupta, Hadzic, Kota, He, Eyzaguirre, Durante, Li, Wu, and Fei-Fei]{hourvideo}
Keshigeyan Chandrasegaran, Agrim Gupta, Lea~M. Hadzic, Taran Kota, Jimming He, Cristóbal Eyzaguirre, Zane Durante, Manling Li, Jiajun Wu, and Li Fei-Fei.
\newblock Hourvideo: 1-hour video-language understanding.
\newblock In \emph{NeurIPS}, 2024.

\bibitem[Chen et~al.(2024{\natexlab{a}})Chen, Wang, Chen, Zhang, Feng, Huang, Jia, and Zhu]{verified}
Houlun Chen, Xin Wang, Hong Chen, Zeyang Zhang, Wei Feng, Bin Huang, Jia Jia, and Wenwu Zhu.
\newblock Verified: A video corpus moment retrieval benchmark for fine-grained video understanding.
\newblock In \emph{NeurIPS}, 2024{\natexlab{a}}.

\bibitem[Chen et~al.(2024{\natexlab{b}})Chen, Wei, Li, Dong, Zhang, Zang, Chen, Duan, Lin, Tang, Yuan, Qiao, Lin, Zhao, and Wang]{sharegpt4video}
Lin Chen, Xilin Wei, Jinsong Li, Xiaoyi Dong, Pan Zhang, Yuhang Zang, Zehui Chen, Haodong Duan, Bin Lin, Zhenyu Tang, Li Yuan, Yu Qiao, Dahua Lin, Feng Zhao, and Jiaqi Wang.
\newblock Sharegpt4video: Improving video understanding and generation with better captions.
\newblock In \emph{NeurIPS}, 2024{\natexlab{b}}.

\bibitem[Chen et~al.(2023)Chen, He, Guo, Zhu, Wang, Tang, and Liu]{valor}
Sihan Chen, Xingjian He, Longteng Guo, Xinxin Zhu, Weining Wang, Jinhui Tang, and Jing Liu.
\newblock Valor: Vision-audio-language omni-perception pretraining model and dataset.
\newblock \emph{arXiv preprint arXiv 2304.08345}, 2023.

\bibitem[Chen et~al.(2024{\natexlab{c}})Chen, Wang, Xue, Zhang, Yang, Li, Shen, Liang, Gu, and Chen]{mhalubench}
Xiang Chen, Chenxi Wang, Yida Xue, Ningyu Zhang, Xiaoyan Yang, Qiang Li, Yue Shen, Lei Liang, Jinjie Gu, and Huajun Chen.
\newblock Unified hallucination detection for multimodal large language models.
\newblock In \emph{ACL}, 2024{\natexlab{c}}.

\bibitem[Cheng et~al.(2024)Cheng, Leng, Zhang, Xin, Li, Chen, Zhu, Zhang, Luo, Zhao, and Bing]{videollama2}
Zesen Cheng, Sicong Leng, Hang Zhang, Yifei Xin, Xin Li, Guanzheng Chen, Yongxin Zhu, Wenqi Zhang, Ziyang Luo, Deli Zhao, and Lidong Bing.
\newblock Videollama 2: Advancing spatial-temporal modeling and audio understanding in video-llms.
\newblock \emph{arXiv preprint arXiv 2406.07476}, 2024.

\bibitem[Cherti et~al.(2023)Cherti, Beaumont, Wightman, Wortsman, Ilharco, Gordon, Schuhmann, Schmidt, and Jitsev]{cherti2023reproducible}
Mehdi Cherti, Romain Beaumont, Ross Wightman, Mitchell Wortsman, Gabriel Ilharco, Cade Gordon, Christoph Schuhmann, Ludwig Schmidt, and Jenia Jitsev.
\newblock Reproducible scaling laws for contrastive language-image learning.
\newblock In \emph{CVPR}, 2023.

\bibitem[Chung et~al.(2024)Chung, Hou, Longpre, Zoph, Tay, Fedus, Li, Wang, Dehghani, Brahma, et~al.]{chung2024scaling}
Hyung~Won Chung, Le Hou, Shayne Longpre, Barret Zoph, Yi Tay, William Fedus, Yunxuan Li, Xuezhi Wang, Mostafa Dehghani, Siddhartha Brahma, et~al.
\newblock Scaling instruction-finetuned language models.
\newblock \emph{JMLR}, 2024.

\bibitem[Dai et~al.(2023)Dai, Li, Li, Tiong, Zhao, Wang, Li, Fung, and Hoi]{instructblip}
Wenliang Dai, Junnan Li, Dongxu Li, Anthony Tiong, Junqi Zhao, Weisheng Wang, Boyang Li, Pascale Fung, and Steven Hoi.
\newblock Instruct{BLIP}: Towards general-purpose vision-language models with instruction tuning.
\newblock In \emph{NeurIPS}, 2023.

\bibitem[Devlin(2018)]{devlin2018bert}
Jacob Devlin.
\newblock Bert: Pre-training of deep bidirectional transformers for language understanding.
\newblock \emph{arXiv preprint arXiv 1810.04805}, 2018.

\bibitem[Driess et~al.(2023)Driess, Xia, Sajjadi, Lynch, Chowdhery, Ichter, Wahid, Tompson, Vuong, Yu, Huang, Chebotar, Sermanet, Duckworth, Levine, Vanhoucke, Hausman, Toussaint, Greff, Zeng, Mordatch, and Florence]{palm-e}
Danny Driess, Fei Xia, Mehdi S.~M. Sajjadi, Corey Lynch, Aakanksha Chowdhery, Brian Ichter, Ayzaan Wahid, Jonathan Tompson, Quan Vuong, Tianhe Yu, Wenlong Huang, Yevgen Chebotar, Pierre Sermanet, Daniel Duckworth, Sergey Levine, Vincent Vanhoucke, Karol Hausman, Marc Toussaint, Klaus Greff, Andy Zeng, Igor Mordatch, and Pete Florence.
\newblock Palm-e: an embodied multimodal language model.
\newblock In \emph{ICML}, 2023.

\bibitem[Fu et~al.(2024)Fu, Dai, Luo, Li, Ren, Zhang, Wang, Zhou, Shen, Zhang, Chen, Li, Lin, Zhao, Li, Xu, Zheng, Chen, Ji, and Sun]{videomme}
Chaoyou Fu, Yuhan Dai, Yongdong Luo, Lei Li, Shuhuai Ren, Renrui Zhang, Zihan Wang, Chenyu Zhou, Yunhang Shen, Mengdan Zhang, Peixian Chen, Yanwei Li, Shaohui Lin, Sirui Zhao, Ke Li, Tong Xu, Xiawu Zheng, Enhong Chen, Rongrong Ji, and Xing Sun.
\newblock Video-mme: The first-ever comprehensive evaluation benchmark of multi-modal llms in video analysis.
\newblock \emph{arXiv preprint arXiv 2405.21075}, 2024.

\bibitem[Gao et~al.(2021)Gao, Yao, and Chen]{simcse}
Tianyu Gao, Xingcheng Yao, and Danqi Chen.
\newblock {S}im{CSE}: Simple contrastive learning of sentence embeddings.
\newblock In \emph{EMNLP}, 2021.

\bibitem[Geigle et~al.(2023)Geigle, Liu, Pfeiffer, and Gurevych]{geigle2023doesfitallcomplementarity}
Gregor Geigle, Chen~Cecilia Liu, Jonas Pfeiffer, and Iryna Gurevych.
\newblock One does not fit all! on the complementarity of vision encoders for vision and language tasks.
\newblock In \emph{RepL4NLP}, 2023.

\bibitem[Guan et~al.(2024)Guan, Liu, Wu, Xian, Li, Liu, Wang, Chen, Huang, Yacoob, Manocha, and Zhou]{hallusionbench}
Tianrui Guan, Fuxiao Liu, Xiyang Wu, Ruiqi Xian, Zongxia Li, Xiaoyu Liu, Xijun Wang, Lichang Chen, Furong Huang, Yaser Yacoob, Dinesh Manocha, and Tianyi Zhou.
\newblock Hallusionbench: An advanced diagnostic suite for entangled language hallucination and visual illusion in large vision-language models.
\newblock In \emph{CVPR}, 2024.

\bibitem[Han et~al.(2024)Han, Lian, Pan, Pi, Zhang, Diao, Lin, and Zhang]{han2024instinctive}
Tianyang Han, Qing Lian, Rui Pan, Renjie Pi, Jipeng Zhang, Shizhe Diao, Yong Lin, and Tong Zhang.
\newblock The instinctive bias: Spurious images lead to hallucination in mllms.
\newblock \emph{arXiv preprint arXiv 2402.03757}, 2024.

\bibitem[Heilbron et~al.(2015)Heilbron, Escorcia, Ghanem, and Niebles]{activitynet}
Fabian~Caba Heilbron, Victor Escorcia, Bernard Ghanem, and Juan~Carlos Niebles.
\newblock Activitynet: A large-scale video benchmark for human activity understanding.
\newblock In \emph{CVPR}, 2015.

\bibitem[Hurst et~al.(2024)Hurst, Lerer, Goucher, Perelman, Ramesh, Clark, Ostrow, Welihinda, Hayes, et~al.]{openai2024gpt4ocard}
Aaron Hurst, Adam Lerer, Adam~P. Goucher, Adam Perelman, Aditya Ramesh, Aidan Clark, AJ Ostrow, Akila Welihinda, Alan Hayes, et~al.
\newblock Gpt-4o system card.
\newblock \emph{arXiv preprint arXiv:2410.21276}, 2024.

\bibitem[Jiang et~al.(2023)Jiang, Liu, Liu, Zhao, Zhang, Gao, Zhang, Li, and Xiong]{jiang2023clip}
Dongsheng Jiang, Yuchen Liu, Songlin Liu, Jin'e Zhao, Hao Zhang, Zhen Gao, Xiaopeng Zhang, Jin Li, and Hongkai Xiong.
\newblock From clip to dino: Visual encoders shout in multi-modal large language models.
\newblock \emph{arXiv preprint arXiv:2310.08825}, 2023.

\bibitem[Jin et~al.(2024)Jin, Takanobu, Zhang, Cao, and Yuan]{chatunivi}
Peng Jin, Ryuichi Takanobu, Wancai Zhang, Xiaochun Cao, and Li Yuan.
\newblock Chat-univi: Unified visual representation empowers large language models with image and video understanding.
\newblock In \emph{CVPR}, 2024.

\bibitem[Koh et~al.(2023)Koh, Salakhutdinov, and Fried]{fromage}
Jing~Yu Koh, Ruslan Salakhutdinov, and Daniel Fried.
\newblock Grounding language models to images for multimodal inputs and outputs.
\newblock In \emph{ICML}, 2023.

\bibitem[Lee et~al.(2024)Lee, Park, Jo, and Seo]{volcano}
Seongyun Lee, Sue~Hyun Park, Yongrae Jo, and Minjoon Seo.
\newblock Volcano: Mitigating multimodal hallucination through self-feedback guided revision.
\newblock In \emph{ACL}, 2024.

\bibitem[Leng et~al.(2024)Leng, Zhang, Chen, Li, Lu, Miao, and Bing]{vcd}
Sicong Leng, Hang Zhang, Guanzheng Chen, Xin Li, Shijian Lu, Chunyan Miao, and Lidong Bing.
\newblock Mitigating object hallucinations in large vision-language models through visual contrastive decoding.
\newblock In \emph{CVPR}, 2024.

\bibitem[Li et~al.(2023{\natexlab{a}})Li, Li, Savarese, and Hoi]{blip2}
Junnan Li, Dongxu Li, Silvio Savarese, and Steven Hoi.
\newblock {BLIP-2}: Bootstrapping language-image pre-training with frozen image encoders and large language models.
\newblock In \emph{ICML}, 2023{\natexlab{a}}.

\bibitem[Li et~al.(2024{\natexlab{a}})Li, Wang, He, Li, Wang, Liu, Wang, Xu, Chen, Luo, Wang, and Qiao]{mvbench}
Kunchang Li, Yali Wang, Yinan He, Yizhuo Li, Yi Wang, Yi Liu, Zun Wang, Jilan Xu, Guo Chen, Ping Luo, Limin Wang, and Yu Qiao.
\newblock Mvbench: A comprehensive multi-modal video understanding benchmark.
\newblock In \emph{CVPR}, 2024{\natexlab{a}}.

\bibitem[Li et~al.(2024{\natexlab{b}})Li, Lei, Gan, Yu, Chen, Pillai, Cheng, Zhou, Wang, Wang, Berg, Bansal, Liu, Wang, and Liu]{value}
Linjie Li, Jie Lei, Zhe Gan, Licheng Yu, Yen-Chun Chen, Rohit Pillai, Yu Cheng, Luowei Zhou, Xin~Eric Wang, William~Yang Wang, Tamara~Lee Berg, Mohit Bansal, Jingjing Liu, Lijuan Wang, and Zicheng Liu.
\newblock Value: A multi-task benchmark for video-and-language understanding evaluation.
\newblock In \emph{NeurIPS}, 2024{\natexlab{b}}.

\bibitem[Li et~al.(2023{\natexlab{b}})Li, Du, Zhou, Wang, Zhao, and Wen]{pope}
Yifan Li, Yifan Du, Kun Zhou, Jinpeng Wang, Xin Zhao, and Ji-Rong Wen.
\newblock Evaluating object hallucination in large vision-language models.
\newblock In \emph{EMNLP}, 2023{\natexlab{b}}.

\bibitem[Li et~al.(2024{\natexlab{c}})Li, Chen, Hu, Wang, Shi, and Zhang]{videovista}
Yunxin Li, Xinyu Chen, Baotian Hu, Longyue Wang, Haoyuan Shi, and Min Zhang.
\newblock Videovista: A versatile benchmark for video understanding and reasoning.
\newblock \emph{arXiv preprint arXiv 2406.11303}, 2024{\natexlab{c}}.

\bibitem[Lin et~al.(2024{\natexlab{a}})Lin, Ye, Zhu, Cui, Ning, Jin, and Yuan]{videollava}
Bin Lin, Yang Ye, Bin Zhu, Jiaxi Cui, Munan Ning, Peng Jin, and Li Yuan.
\newblock Video-llava: Learning united visual representation by alignment before projection.
\newblock In \emph{EMNLP}, 2024{\natexlab{a}}.

\bibitem[Lin et~al.(2024{\natexlab{b}})Lin, Yin, Ping, Lu, Molchanov, Tao, Mao, Kautz, Shoeybi, and Han]{vila}
Ji Lin, Hongxu Yin, Wei Ping, Yao Lu, Pavlo Molchanov, Andrew Tao, Huizi Mao, Jan Kautz, Mohammad Shoeybi, and Song Han.
\newblock Vila: On pre-training for visual language models.
\newblock In \emph{CVPR}, 2024{\natexlab{b}}.

\bibitem[Liu et~al.(2024{\natexlab{a}})Liu, Lin, Li, Wang, Yacoob, and Wang]{lrv-instruction}
Fuxiao Liu, Kevin Lin, Linjie Li, Jianfeng Wang, Yaser Yacoob, and Lijuan Wang.
\newblock Mitigating hallucination in large multi-modal models via robust instruction tuning.
\newblock In \emph{ICLR}, 2024{\natexlab{a}}.

\bibitem[Liu et~al.(2023)Liu, Li, Wu, and Lee]{llava}
Haotian Liu, Chunyuan Li, Qingyang Wu, and Yong~Jae Lee.
\newblock Visual instruction tuning.
\newblock In \emph{NeurIPS}, 2023.

\bibitem[Liu et~al.(2024{\natexlab{b}})Liu, Li, Li, and Lee]{liu2024improved}
Haotian Liu, Chunyuan Li, Yuheng Li, and Yong~Jae Lee.
\newblock Improved baselines with visual instruction tuning.
\newblock In \emph{CVPR}, 2024{\natexlab{b}}.

\bibitem[Liu et~al.(2024{\natexlab{c}})Liu, Fu, Xie, Xie, Sun, Lian, Kang, and Li]{phd}
Jiazhen Liu, Yuhan Fu, Ruobing Xie, Runquan Xie, Xingwu Sun, Fengzong Lian, Zhanhui Kang, and Xirong Li.
\newblock Phd: A prompted visual hallucination evaluation dataset.
\newblock \emph{arXiv preprint arXiv 2403.11116}, 2024{\natexlab{c}}.

\bibitem[Liu et~al.(2024{\natexlab{d}})Liu, Duan, Zhang, Li, Zhang, Zhao, Yuan, Wang, He, Liu, Chen, and Lin]{mmbench}
Yuan Liu, Haodong Duan, Yuanhan Zhang, Bo Li, Songyang Zhang, Wangbo Zhao, Yike Yuan, Jiaqi Wang, Conghui He, Ziwei Liu, Kai Chen, and Dahua Lin.
\newblock Mmbench: Is your multi-modal model an all-around player?
\newblock In \emph{ECCV}, 2024{\natexlab{d}}.

\bibitem[Maaz et~al.(2024)Maaz, Rasheed, Khan, and Khan]{videochatgpt}
Muhammad Maaz, Hanoona Rasheed, Salman Khan, and Fahad Khan.
\newblock Video-{C}hat{GPT}: Towards detailed video understanding via large vision and language models.
\newblock In \emph{ACL}, 2024.

\bibitem[Mangalam et~al.(2023)Mangalam, Akshulakov, and Malik]{mangalam2023egoschema}
Karttikeya Mangalam, Raiymbek Akshulakov, and Jitendra Malik.
\newblock Egoschema: A diagnostic benchmark for very long-form video language understanding.
\newblock In \emph{NeurIPS}, 2023.

\bibitem[Naeem et~al.(2024)Naeem, Xian, Zhai, Hoyer, Van~Gool, and Tombari]{naeem2025silc}
Muhammad~Ferjad Naeem, Yongqin Xian, Xiaohua Zhai, Lukas Hoyer, Luc Van~Gool, and Federico Tombari.
\newblock Silc: Improving vision language pretraining with self-distillation.
\newblock In \emph{ECCV}, 2024.

\bibitem[Nguyen et~al.(2024{\natexlab{a}})Nguyen, Yamagishi, and Echizen]{nguyen2024exploring}
Huy~H Nguyen, Junichi Yamagishi, and Isao Echizen.
\newblock Exploring self-supervised vision transformers for deepfake detection: A comparative analysis.
\newblock \emph{arXiv preprint arXiv:2405.00355}, 2024{\natexlab{a}}.

\bibitem[Nguyen et~al.(2024{\natexlab{b}})Nguyen, Bi, Vosoughi, Tian, Fazli, and Xu]{oscar}
Nguyen Nguyen, Jing Bi, Ali Vosoughi, Yapeng Tian, Pooyan Fazli, and Chenliang Xu.
\newblock Oscar: Object state captioning and state change representation.
\newblock In \emph{NAACL}, 2024{\natexlab{b}}.

\bibitem[Oquab et~al.(2024)Oquab, Darcet, Moutakanni, Vo, Szafraniec, Khalidov, Fernandez, HAZIZA, Massa, El-Nouby, Assran, Ballas, Galuba, Howes, Huang, Li, Misra, Rabbat, Sharma, Synnaeve, Xu, Jegou, Mairal, Labatut, Joulin, and Bojanowski]{dinov2}
Maxime Oquab, Timoth{\'e}e Darcet, Th{\'e}o Moutakanni, Huy~V. Vo, Marc Szafraniec, Vasil Khalidov, Pierre Fernandez, Daniel HAZIZA, Francisco Massa, Alaaeldin El-Nouby, Mido Assran, Nicolas Ballas, Wojciech Galuba, Russell Howes, Po-Yao Huang, Shang-Wen Li, Ishan Misra, Michael Rabbat, Vasu Sharma, Gabriel Synnaeve, Hu Xu, Herve Jegou, Julien Mairal, Patrick Labatut, Armand Joulin, and Piotr Bojanowski.
\newblock {DINO}v2: Learning robust visual features without supervision.
\newblock \emph{TMLR}, 2024.

\bibitem[Peng et~al.(2023)Peng, Li, He, Galley, and Gao]{vicuna}
Baolin Peng, Chunyuan Li, Pengcheng He, Michel Galley, and Jianfeng Gao.
\newblock Instruction tuning with gpt-4.
\newblock \emph{arXiv preprint arXiv 2304.03277}, 2023.

\bibitem[Radford et~al.(2021)Radford, Kim, Hallacy, Ramesh, Goh, Agarwal, Sastry, Askell, Mishkin, Clark, Krueger, and Sutskever]{clip}
Alec Radford, Jong~Wook Kim, Chris Hallacy, Aditya Ramesh, Gabriel Goh, Sandhini Agarwal, Girish Sastry, Amanda Askell, Pamela Mishkin, Jack Clark, Gretchen Krueger, and Ilya Sutskever.
\newblock Learning transferable visual models from natural language supervision.
\newblock \emph{PMLR}, 2021.

\bibitem[Raffel et~al.(2020)Raffel, Shazeer, Roberts, Lee, Narang, Matena, Zhou, Li, and Liu]{t5}
Colin Raffel, Noam Shazeer, Adam Roberts, Katherine Lee, Sharan Narang, Michael Matena, Yanqi Zhou, Wei Li, and Peter~J Liu.
\newblock Exploring the limits of transfer learning with a unified text-to-text transformer.
\newblock \emph{JMLR}, 2020.

\bibitem[Rodriguez-Opazo et~al.(2023)Rodriguez-Opazo, Marrese-Taylor, Abbasnejad, Damirchi, Jara, Bravo-Marquez, and van~den Hengel]{rodriguezopazo2023unveilingbackboneeffectsclip}
Cristian Rodriguez-Opazo, Edison Marrese-Taylor, Ehsan Abbasnejad, Hamed Damirchi, Ignacio~M. Jara, Felipe Bravo-Marquez, and Anton van~den Hengel.
\newblock Unveiling backbone effects in clip: Exploring representational synergies and variances.
\newblock \emph{arXiv preprint arXiv:2312.14400}, 2023.

\bibitem[Rohrbach et~al.(2018)Rohrbach, Hendricks, Burns, Darrell, and Saenko]{rohrbach2018object}
Anna Rohrbach, Lisa~Anne Hendricks, Kaylee Burns, Trevor Darrell, and Kate Saenko.
\newblock Object hallucination in image captioning.
\newblock \emph{arXiv preprint arXiv 1809.02156}, 2018.

\bibitem[Sun et~al.(2023{\natexlab{a}})Sun, Fang, Wu, Wang, and Cao]{evaclip}
Quan Sun, Yuxin Fang, Ledell Wu, Xinlong Wang, and Yue Cao.
\newblock Eva-clip: Improved training techniques for clip at scale.
\newblock \emph{arXiv preprint arXiv 2303.15389}, 2023{\natexlab{a}}.

\bibitem[Sun et~al.(2023{\natexlab{b}})Sun, Shen, Cao, Liu, Li, Shen, Gan, Gui, Wang, Yang, Keutzer, and Darrell]{mmhal_bench}
Zhiqing Sun, Sheng Shen, Shengcao Cao, Haotian Liu, Chunyuan Li, Yikang Shen, Chuang Gan, Liang-Yan Gui, Yu-Xiong Wang, Yiming Yang, Kurt Keutzer, and Trevor Darrell.
\newblock Aligning large multimodal models with factually augmented rlhf.
\newblock \emph{arXiv preprint arXiv 2309.14525}, 2023{\natexlab{b}}.

\bibitem[Tang et~al.(2023)Tang, Bi, Xu, Song, Liang, Wang, Zhang, An, Lin, Zhu, Vosoughi, Huang, Zhang, Liu, Feng, Zheng, Zhang, Luo, Luo, and Xu]{mllm_survey}
Yunlong Tang, Jing Bi, Siting Xu, Luchuan Song, Susan Liang, Teng Wang, Daoan Zhang, Jie An, Jingyang Lin, Rongyi Zhu, Ali Vosoughi, Chao Huang, Zeliang Zhang, Pinxin Liu, Mingqian Feng, Feng Zheng, Jianguo Zhang, Ping Luo, Jiebo Luo, and Chenliang Xu.
\newblock Video understanding with large language models: A survey.
\newblock \emph{arXiv preprint arXiv 2312.17432}, 2023.

\bibitem[Team et~al.(2024)Team, Georgiev, Lei, Burnell, Bai, Gulati, Tanzer, Vincent, Pan, Wang, et~al.]{team2024gemini}
Gemini Team, Petko Georgiev, Ving~Ian Lei, Ryan Burnell, Libin Bai, Anmol Gulati, Garrett Tanzer, Damien Vincent, Zhufeng Pan, Shibo Wang, et~al.
\newblock Gemini 1.5: Unlocking multimodal understanding across millions of tokens of context.
\newblock \emph{arXiv preprint arXiv 2403.05530}, 2024.

\bibitem[Tong et~al.(2024{\natexlab{a}})Tong, II, Wu, Woo, IYER, Akula, Yang, Yang, Middepogu, Wang, Pan, Fergus, LeCun, and Xie]{cambrian}
Shengbang Tong, Ellis L~Brown II, Penghao Wu, Sanghyun Woo, ADITHYA~JAIRAM IYER, Sai~Charitha Akula, Shusheng Yang, Jihan Yang, Manoj Middepogu, Ziteng Wang, Xichen Pan, Rob Fergus, Yann LeCun, and Saining Xie.
\newblock Cambrian-1: A fully open, vision-centric exploration of multimodal {LLM}s.
\newblock In \emph{NeurIPS}, 2024{\natexlab{a}}.

\bibitem[Tong et~al.(2024{\natexlab{b}})Tong, Liu, Zhai, Ma, LeCun, and Xie]{mmvp}
Shengbang Tong, Zhuang Liu, Yuexiang Zhai, Yi Ma, Yann LeCun, and Saining Xie.
\newblock Eyes wide shut? exploring the visual shortcomings of multimodal llms.
\newblock In \emph{CVPR}, 2024{\natexlab{b}}.

\bibitem[Touvron et~al.(2023)Touvron, Lavril, Izacard, Martinet, Lachaux, Lacroix, Rozière, Goyal, Hambro, Azhar, Rodriguez, Joulin, Grave, and Lample]{llama}
Hugo Touvron, Thibaut Lavril, Gautier Izacard, Xavier Martinet, Marie-Anne Lachaux, Timothée Lacroix, Baptiste Rozière, Naman Goyal, Eric Hambro, Faisal Azhar, Aurelien Rodriguez, Armand Joulin, Edouard Grave, and Guillaume Lample.
\newblock Llama: Open and efficient foundation language models.
\newblock \emph{arXiv preprint arXiv 2302.13971}, 2023.

\bibitem[Wang et~al.(2023)Wang, Wang, Xu, Zhang, Gu, Jia, Wang, Xu, Yan, Zhang, and Sang]{amber}
Junyang Wang, Yuhang Wang, Guohai Xu, Jing Zhang, Yukai Gu, Haitao Jia, Jiaqi Wang, Haiyang Xu, Ming Yan, Ji Zhang, and Jitao Sang.
\newblock Amber: An llm-free multi-dimensional benchmark for mllms hallucination evaluation.
\newblock \emph{arXiv preprint arXiv 2311.07397}, 2023.

\bibitem[Wang et~al.(2024)Wang, Wang, Zhao, Xie, and Zheng]{videohallucer}
Yuxuan Wang, Yueqian Wang, Dongyan Zhao, Cihang Xie, and Zilong Zheng.
\newblock Videohallucer: Evaluating intrinsic and extrinsic hallucinations in large video-language models.
\newblock \emph{arXiv preprint arXiv 2406.16338}, 2024.

\bibitem[Wei et~al.(2021)Wei, Bosma, Zhao, Guu, Yu, Lester, Du, Dai, and Le]{wei2021finetuned}
Jason Wei, Maarten Bosma, Vincent~Y Zhao, Kelvin Guu, Adams~Wei Yu, Brian Lester, Nan Du, Andrew~M Dai, and Quoc~V Le.
\newblock Finetuned language models are zero-shot learners.
\newblock \emph{arXiv preprint arXiv 2109.01652}, 2021.

\bibitem[Wu et~al.(2024)Wu, Li, Chen, and Li]{longvideobench}
Haoning Wu, Dongxu Li, Bei Chen, and Junnan Li.
\newblock Longvideobench: A benchmark for long-context interleaved video-language understanding.
\newblock In \emph{NeurIPS}, 2024.

\bibitem[Wysoczanska et~al.(2024)Wysoczanska, Sim{\'e}oni, Ramamonjisoa, Bursuc, Trzcinski, and P{\'e}rez]{wysoczanska2024clip}
Monika Wysoczanska, Oriane Sim{\'e}oni, Micha{\"e}l Ramamonjisoa, Andrei Bursuc, Tomasz Trzcinski, and Patrick P{\'e}rez.
\newblock Clip-dinoiser: Teaching clip a few dino tricks for open-vocabulary semantic segmentation.
\newblock In \emph{ECCV}, 2024.

\bibitem[Xu et~al.(2021)Xu, Ghosh, Huang, Okhonko, Aghajanyan, Metze, Zettlemoyer, and Feichtenhofer]{videoclip}
Hu Xu, Gargi Ghosh, Po-Yao Huang, Dmytro Okhonko, Armen Aghajanyan, Florian Metze, Luke Zettlemoyer, and Christoph Feichtenhofer.
\newblock Videoclip: Contrastive pre-training for zero-shot video-text understanding.
\newblock In \emph{EMNLP}, 2021.

\bibitem[Xu et~al.(2024)Xu, Zhao, Zhou, Lin, Ng, and Feng]{pllava}
Lin Xu, Yilin Zhao, Daquan Zhou, Zhijie Lin, See~Kiong Ng, and Jiashi Feng.
\newblock Pllava : Parameter-free llava extension from images to videos for video dense captioning.
\newblock \emph{arXiv preprint arXiv 2404.16994}, 2024.

\bibitem[Yang et~al.(2024)Yang, Huang, Lu, Han, Zhang, Gao, Hu, and Zhao]{vript}
Dongjie Yang, Suyuan Huang, Chengqiang Lu, Xiaodong Han, Haoxin Zhang, Yan Gao, Yao Hu, and Hai Zhao.
\newblock Vript: A video is worth thousands of words.
\newblock In \emph{NeurIPS}, 2024.

\bibitem[Yin et~al.(2023)Yin, Fu, Zhao, Xu, Wang, Sui, Shen, Li, Sun, and Chen]{woodpecker}
Shukang Yin, Chaoyou Fu, Sirui Zhao, Tong Xu, Hao Wang, Dianbo Sui, Yunhang Shen, Ke Li, Xing Sun, and Enhong Chen.
\newblock Woodpecker: Hallucination correction for multimodal large language models.
\newblock \emph{arXiv preprint arXiv 2310.16045}, 2023.

\bibitem[Yu et~al.(2024{\natexlab{a}})Yu, Li, Wei, Pang, Ye, Qin, Tang, Tian, and Zhuang]{hallucidoctor}
Qifan Yu, Juncheng Li, Longhui Wei, Liang Pang, Wentao Ye, Bosheng Qin, Siliang Tang, Qi Tian, and Yueting Zhuang.
\newblock Hallucidoctor: Mitigating hallucinatory toxicity in visual instruction data.
\newblock In \emph{CVPR}, 2024{\natexlab{a}}.

\bibitem[Yu et~al.(2024{\natexlab{b}})Yu, Yao, Zhang, He, Han, Cui, Hu, Liu, Zheng, Sun, et~al.]{yu2024rlhf}
Tianyu Yu, Yuan Yao, Haoye Zhang, Taiwen He, Yifeng Han, Ganqu Cui, Jinyi Hu, Zhiyuan Liu, Hai-Tao Zheng, Maosong Sun, et~al.
\newblock Rlhf-v: Towards trustworthy mllms via behavior alignment from fine-grained correctional human feedback.
\newblock In \emph{CVPR}, 2024{\natexlab{b}}.

\bibitem[Zhai et~al.(2023)Zhai, Mustafa, Kolesnikov, and Beyer]{siglip}
Xiaohua Zhai, Basil Mustafa, Alexander Kolesnikov, and Lucas Beyer.
\newblock Sigmoid loss for language image pre-training.
\newblock In \emph{ICCV}, 2023.

\bibitem[Zhang et~al.(2025)Zhang, Li, Cheng, Hu, Yuan, Chen, Leng, Jiang, Zhang, Li, Jin, Zhang, Wang, Bing, and Zhao]{videollama3}
Boqiang Zhang, Kehan Li, Zesen Cheng, Zhiqiang Hu, Yuqian Yuan, Guanzheng Chen, Sicong Leng, Yuming Jiang, Hang Zhang, Xin Li, Peng Jin, Wenqi Zhang, Fan Wang, Lidong Bing, and Deli Zhao.
\newblock Videollama 3: Frontier multimodal foundation models for image and video understanding.
\newblock \emph{arXiv preprint arXiv 2501.13106}, 2025.

\bibitem[Zhang et~al.(2023)Zhang, Li, Liu, Zhang, Su, Zhu, Ni, and Shum]{dino}
Hao Zhang, Feng Li, Shilong Liu, Lei Zhang, Hang Su, Jun Zhu, Lionel Ni, and Heung-Yeung Shum.
\newblock {DINO}: {DETR} with improved denoising anchor boxes for end-to-end object detection.
\newblock In \emph{ICLR}, 2023.

\bibitem[Zhang et~al.(2024{\natexlab{a}})Zhang, Jiao, Chen, Chen, and Jiang]{eventhallusion}
Jiacheng Zhang, Yang Jiao, Shaoxiang Chen, Jingjing Chen, and Yu-Gang Jiang.
\newblock Eventhallusion: Diagnosing event hallucinations in video llms.
\newblock \emph{arXiv preprint arXiv 2409.16597}, 2024{\natexlab{a}}.

\bibitem[Zhang et~al.(2024{\natexlab{b}})Zhang, Li, Liu, Lee, Gui, Fu, Feng, Liu, and Li]{llavanextvideo}
Yuanhan Zhang, Bo Li, haotian Liu, Yong~jae Lee, Liangke Gui, Di Fu, Jiashi Feng, Ziwei Liu, and Chunyuan Li.
\newblock {LLaVA-NeXT}: A strong zero-shot video understanding model, 2024{\natexlab{b}}.

\bibitem[Zhao et~al.(2023)Zhao, Wang, Ouyang, Dong, Wang, and He]{ha_dpo}
Zhiyuan Zhao, Bin Wang, Linke Ouyang, Xiaoyi Dong, Jiaqi Wang, and Conghui He.
\newblock Beyond hallucinations: Enhancing lvlms through hallucination-aware direct preference optimization.
\newblock \emph{arXiv preprint arXiv 2311.16839}, 2023.

\bibitem[Zhou et~al.(2018)Zhou, Xu, and Corso]{youcook2}
Luowei Zhou, Chenliang Xu, and Jason~J. Corso.
\newblock Towards automatic learning of procedures from web instructional videos.
\newblock In \emph{AAAI}, 2018.

\bibitem[Zhou et~al.(2024)Zhou, Cui, Rafailov, Finn, and Yao]{zhou2024aligning}
Yiyang Zhou, Chenhang Cui, Rafael Rafailov, Chelsea Finn, and Huaxiu Yao.
\newblock Aligning modalities in vision large language models via preference fine-tuning.
\newblock \emph{arXiv preprint arXiv 2402.11411}, 2024.

\bibitem[Zhu et~al.(2024{\natexlab{a}})Zhu, Lin, Ning, Yan, Cui, HongFa, Pang, Jiang, Zhang, Li, Zhang, Li, Liu, and Yuan]{languagebind}
Bin Zhu, Bin Lin, Munan Ning, Yang Yan, Jiaxi Cui, WANG HongFa, Yatian Pang, Wenhao Jiang, Junwu Zhang, Zongwei Li, Cai~Wan Zhang, Zhifeng Li, Wei Liu, and Li Yuan.
\newblock Languagebind: Extending video-language pretraining to n-modality by language-based semantic alignment.
\newblock In \emph{ICLR}, 2024{\natexlab{a}}.

\bibitem[Zhu et~al.(2024{\natexlab{b}})Zhu, Chen, Shen, Li, and Elhoseiny]{minigpt}
Deyao Zhu, Jun Chen, Xiaoqian Shen, Xiang Li, and Mohamed Elhoseiny.
\newblock Mini{GPT}-4: Enhancing vision-language understanding with advanced large language models.
\newblock In \emph{ICLR}, 2024{\natexlab{b}}.

\end{thebibliography}
